\documentclass[lettersize,journal]{IEEEtran}
\usepackage[utf8]{inputenc} 
\usepackage{graphicx} 
\usepackage{xcolor}
\usepackage[normalem]{ulem} 
 \usepackage{xr}

\graphicspath{{./visuals/}}
\title{K-VARK: Kernelized Variance-Aware Residual Kalman Filter for Sensorless Force Estimation in Collaborative Robots}
\author{
Oğuzhan~Akbıyık, Naseem~Alhousani, ~Fares~J.~Abu-Dakka,~\IEEEmembership{Senior~Member,~IEEE}%

\thanks{
(Corresponding author: F. J. Abu-Dakka)
}%
\thanks{
O. Akbıyık is with TU Wien, 1040 Vienna, Austria (e-mail: oguzhan.akbiyik@tuwien.ac.at; oguzhan-akbiyik@hotmail.com).
This work was partly conducted while he was with MCFLY Robot Technologies, Istanbul 34475, Turkey.
}
\thanks{N. Alhousani is with MCFLY Robot Technologies, Istanbul 34475, Turkey (e-mail: naseem.alhousani@mcflyrobot.com). 
}
\thanks{
F. J. Abu-Dakka is with Mechanical Engineering Program, New York University Abu Dhabi, Abu Dhabi, United Arab Emirates (e-mail: fa2656@nyu.edu).
 ORCID: 0000-0001-9062-9416
}%
\thanks{
This work was partially supported by the NYUAD Center for Artificial Intelligence and Robotics (CAIR), funded by Tamkeen under the NYUAD Research Institute Award CG010.
}
\thanks{
The authors would like to thank MCFLY Robot Technologies for providing access to the Orion 5 collaborative robot platform and supporting the data collection experiments.
}
}
\usepackage{amsmath,amssymb,mathtools,bm}  
\usepackage{algorithm}
\usepackage[noend]{algpseudocode}
\algrenewcommand\algorithmicrequire{\textbf{Input:}}
\algrenewcommand\algorithmicensure{\textbf{Output:}}
\usepackage{mathrsfs}  

\usepackage{graphicx}                      
\usepackage{siunitx} 
\usepackage{booktabs}                      
\usepackage{enumitem} 
\usepackage{tikz}
\usetikzlibrary{arrows.meta,positioning,shapes.geometric,calc}
\usepackage{algorithm}                     
\usepackage[noend]{algpseudocode}          
\usepackage{hyperref}                      

\usepackage[nolist]{acronym} 

\newacro{kvark}[K-VARK]{Kernelized Variance-Aware Residual Kalman filter}
\newacro{gmm}[GMM]{Gaussian Mixture Models}
\newacro{gmr}[GMR]{Gaussian Mixture Regression}
\newacro{em}[EM]{Expectation-Minimization}
\newacro{kmp}[KMP]{Kernelized Movement Primitives}
\newacro{gp}[GP]{Gaussian Process}
\newacro{gpr}[GPR]{Gaussian Process Regression}
\newacro{gpadkf}[GPADKF]{Gaussian Process Adaptive Disturbance Kalman Filter}
\newacro{kf}[KF]{Kalman Filter}
\newacro{adkf}[ADKF]{Adaptive Disturbance Kalman Filter}
\newacro{nn}[NN]{Neural Networks}
\newacro{dob}[DOB]{Disturbance Observer}
\newacro{ndo}[NDO]{Nonlinear Disturbance Observer}
\newacro{gmo}[GMO]{Generalized Momentum Observer}
\newacro{ekf}[EKF]{Extended Kalman Filter}
\newacro{dkf}[DKF]{Disturbance Kalman Filter}
\newacro{kl}[KL]{Kullback–Leibler}
\newacro{akf}[AKF]{Adaptive Kalman Filter}
\newacro{vb}[VB]{Variational Bayesian}
\newacro{rmse}[RMSE]{Root Mean Squared Error}
\newacro{ga}[GA]{Genetic Algorithms}
\newacro{ard}[ARD]{Automatic Relevance Determination}
\newacro{nis}[NIS]{ Normalized Innovation Squared}
\newacro{ros}[ROS]{Robot Operating System}
\newacro{nnadkf}[NNADKF]{Neural Network Based Kalman Filter}
\input{macros}
\newtheorem{remark}{Remark}
\begin{document}

\maketitle
\begin{abstract}
    Reliable estimation of contact forces is crucial for ensuring safe and precise interaction of robots with unstructured environments. However, accurate sensorless force estimation remains challenging due to inherent modeling errors, complex residual dynamics and friction. To address this challenge, we propose K-VARK (Kernelized Variance-Aware Residual Kalman filter), a novel approach that integrates a kernelized, probabilistic model of joint residual torques into an adaptive Kalman filter framework. Through Kernelized Movement Primitives trained on optimized excitation trajectories, K-VARK captures both the predictive mean and input-dependent heteroscedastic variance of residual torques, reflecting data variability and distance-to-training effects. These statistics inform a variance-aware virtual measurement update by augmenting the measurement noise covariance, while the process noise covariance adapts online via variational Bayesian optimization to handle dynamic disturbances. Experimental validation on a 6-DoF collaborative manipulator demonstrates that K-VARK achieves over 20\% reduction in RMSE compared to state-of-the-art sensorless force estimation methods, yielding robust and accurate external force/torque estimation suitable for tasks such as polishing and assembly.
\end{abstract}
\def\abstractname{Note to Practitioners}
\begin{abstract}
In many industrial and collaborative robotics applications, such as polishing, assembly, and surface finishing, accurate force estimation is essential for safety and performance, but is often limited by the cost and fragility of physical force/torque sensors. This work introduces K-VARK (Kernelized Variance-Aware Residual Kalman Filter), a practical method for sensorless force estimation that combines probabilistic learning with adaptive filtering. K-VARK models both the mean and variability of \changeOLD{unmodeled} residual torques, allowing the filter to dynamically adjust its confidence based on operating conditions. Experiments on a 6-DoF collaborative robot show over 20 percent improvement in estimation accuracy compared to existing approaches, with real-time computation suitable for deployment in industrial controllers. The method enables cost-effective, robust external force estimation for contact-rich tasks without additional sensors, enhancing reliability in automation and human-robot interaction. Future extensions will target broader robot types and improved handling of low-velocity and temperature-dependent friction effects. 
\end{abstract}
\begin{IEEEkeywords}
Collaborative robots, kernelized movement primitives, probabilistic modeling, residual Kalman filtering, force estimation.
\end{IEEEkeywords}
\begin{figure}[]
  \centering  
      \def\svgwidth{1\linewidth}
      \fontsize{9}{9}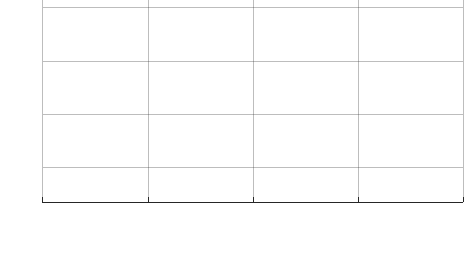%
  \caption{ Comparison between \ac{kf} behaviors using \ac{gp} or \ac{kmp} for process \changeOLD{modeling}}
\label{fig:frontpage}
\end{figure}
\section{Introduction}
\IEEEPARstart{R}{obotics} is revolutionizing the manufacturing industry, addressing critical challenges such as labor shortages and the growing demand for efficiency and precision~\cite{russmann2015industry}. Whereas industrial robots were traditionally deployed for repetitive tasks, modern applications increasingly require robots to perform complex operations that involve dynamic interactions with their environments~\cite{keshvarparast2024collaborative}. In many applications, such as polishing, deburring, and assembly, robots are expected not only to maintain physical contact but also to regulate interaction forces with high accuracy, where task performance depends on the quality and control of applied forces.
 
To satisfy these demanding requirements, many industrial robots are equipped with dedicated force-torque sensors located at the end effector or integrated joint torque sensors~\cite{ke2023review}. Although these sensors improve performance, they also add considerable cost and complexity to robotic systems. This motivates a growing interest in \emph{sensorless force estimation} methods~\cite{liu2021sensorless,ibari2024collision}, which infer contact forces using only internal robot signals and dynamic modeling, thus reducing the dependence on costly force sensing.

Sensorless force estimation typically employs force observers using identified models of the robot's dynamics~\cite{han2021toward}. However, these approaches are vulnerable to inaccuracies arising from sensor noise, parameter drift, incomplete or biased excitation in training data, and model mismatches~\cite{madsen2020comprehensive}. Such imperfections propagate to force observers and degrade the reliability and generalizability of force estimates. Among the most challenging factors is joint friction, whose nonlinear and often poorly characterized behavior introduces complex residual dynamics~\cite{kermani2007friction}.

A critical limitation often overlooked in existing friction estimation techniques is that the modeling data do not consist solely of frictional effects; what is assumed to be ``pure friction" also contains residuals from other identification errors and modeling inaccuracies~\cite{huang2025adaptive}. This combined effect, termed \emph{residual torque}~\cite{flacco2011residual}, leads to unreliable force estimation if treated strictly as friction. Although some approaches probabilistically model residual torques~\cite{giacomuzzo2024black}, they commonly ignore the intrinsic variance caused by these heterogeneous error sources. Accurately capturing this variance is vital to understanding system uncertainty and improving observer robustness.

\ac{kmp}~\cite{huang2019kernelized}, a kernel-based probabilistic model originally developed for imitation learning, jointly captures the mean and input-dependent (heteroscedastic) variance of residual dynamics---the property the methods above overlook. Figure \ref{fig:frontpage} shows the distinct behavior of \acp{kf} using \ac{gp} and \ac{kmp} residual models in two data regimes: region~A with densely sampled but noisy effects and region~B with sparsely populated input data. The \ac{kmp}-based filter effectively inflates model noise in region A, thus adaptively increasing the reliance on the measurements and mitigating overconfidence. In contrast, the \ac{gp}-based filter underestimates variance in the same region, resulting in oscillations due to overtrusting the noisy modeled process. Importantly, in sparse region B, both filters exhibit similar uncertainty quantification, highlighting \ac{kmp}’s ability to reliably represent epistemic uncertainty without compromising its variance adaptation.

To address these challenges, we propose \ac{kvark}, a novel external force observer that integrates variance-aware residual torque modeling---via \ac{kmp}, initialized from \ac{gmm}---with adaptive filtering. By jointly exploiting residual mean, uncertainty, and heteroscedastic variance within an adaptive Kalman filter measurement model, and by adapting force-dynamics covariance online through variational Bayesian optimization~\cite{sarkka2009recursive}, \ac{kvark} achieves high estimation accuracy while maintaining computational efficiency.

Our main contributions are as follows:
\begin{itemize}
    \item We introduce \ac{kvark}, a novel sensorless external force observer that couples probabilistic \ac{kmp}-based residual torque modeling with adaptive Kalman filtering to robustly estimate external forces on robotic manipulators without force-torque sensors.

    \item We validate the proposed method through extensive real-world experiments on a 6-DoF collaborative robot, demonstrating improved accuracy and computational efficiency compared to state-of-the-art sensorless force estimation techniques.
\end{itemize}
\noindent The remainder of this paper is organized as follows: 
Section~\ref{sec:related_work} reviews related work on sensorless force estimation and residual torque modeling. 
Section~\ref{sec:background} provides the necessary background on robot dynamics, momentum-based models, and the Kalman filtering framework. 
Section~\ref{sec:proposed method} presents the proposed \ac{kvark} method, including residual torque modeling with \ac{kmp} and its integration into an adaptive Kalman filter. 
Section~\ref{sec:experiments} reports the experimental setup and evaluation results on a 6-DoF collaborative robot. 
Section~\ref{sec:Discussion} discusses design choices, feature selection, and limitations of the approach. 
Finally, Section~\ref{sec:conclusion} concludes the paper with future research directions. 

\section{Related Work}\label{sec:related_work}

Early research on sensorless force estimation was mainly based on analytical models and observer-based techniques. A conventional strategy compared the commanded actuator torque with that predicted by a nominal model, interpreting discrepancies as external forces. De Luca and colleagues noted that acceleration readings are noisy and instead derived residual torques from the generalized momentum model~\cite{de2005sensorless, de2006collision}. Although intuitive, this approach requires carefully tuned thresholds and is sensitive to model uncertainties and actuator noise: slight variations in friction or inertial parameters can produce residual torques that may be misclassified as physical contact, thereby complicating robust detection.

To address these limitations, research has evolved along two fronts: improved dynamical modeling and advanced observer design. In dynamical modeling, residual torques---arising from unmodeled dynamics, uncertain parameters, and nonlinear friction---have received growing attention. Classical models such as the Coulomb model~\cite{pennestri2016review} (assuming constant friction magnitude) and the LuGre model~\cite{johanastrom2008revisiting} (capturing microslip dynamics) provide foundational baselines, but modern manufacturing tasks increasingly demand more accurate and flexible representations. Consequently, learning-based and data-driven strategies have grown in prominence, including neural-network-based friction modeling~\cite{vantilborgh2024probabilistic, scholl2024learning} and genetic algorithms for complex or non-smooth joint behaviors~\cite{tu2019modeling}. Despite their empirical successes, fully learning-based methods often suffer from limited interpretability, higher computational cost, and poor sample efficiency---issues that restrict their use in real-time industrial scenarios.

\renewcommand{\arraystretch}{1.2}
\begin{table*}[ht]
\centering
\footnotesize
\caption{Comparison of residual torque modeling and force estimation approaches}
\label{tab:comparison}
\begin{tabular}{p{4.2cm} p{2.9cm} p{3.7cm} p{1.8cm} p{3.2cm}}
\toprule
\textbf{Method / Reference} & \textbf{Friction/Residual Modeling} & \textbf{Uncertainty Source/Handling} & \textbf{Real-Time Suitability} & \textbf{External Force Estimation} \\
\midrule
Threshold-based residuals \cite{de2005sensorless, de2006collision} & Nominal dynamics only & None & High & Basic (threshold-dependent) \\
Neural network models \cite{vantilborgh2024probabilistic, scholl2024learning} & Nonlinear, learned & Probabilistic model noise \cite{vantilborgh2024probabilistic}, None/Implicit \cite{scholl2024learning} & Moderate--Low & Not Demonstrated \cite{vantilborgh2024probabilistic} \newline Yes \cite{scholl2024learning} \\
Semi-parametric \ac{nn} + model \cite{peng2020neural, hu2017contact} & Combined analytic + learned & None \cite{peng2020neural}, KF covariance propagation \cite{hu2017contact} & Moderate & Yes \\
Semi-parametric \ac{gp} \cite{wu2012semi} & Combined analytic + learned & Predictive variance + noise model & Moderate & Not Demonstrated \\
\ac{gpadkf} \cite{wei2023contact} & Learned residuals (\ac{gp} mean) & \ac{gp} covariance + \ac{vb} & Moderate & Yes \\
\textbf{Our method (\ac{kvark})} & Learned residuals (\ac{kmp} mean + variability) & Heteroscedastic variance learning + empirical adaptation + \ac{vb} & High & Yes \\
\bottomrule
\end{tabular}
\end{table*}

Semi-parametric approaches, which combine analytical models with flexible learning-based components, balance transparency and adaptability. For example, Peng et al.~\cite{peng2020neural} integrated a radial basis neural network within an admittance control framework, where the network compensates for model uncertainties while an inverse-dynamics force observer estimates external forces. Similarly, Hu and Xiong~\cite{hu2017contact} proposed a hybrid scheme coupling a rigid-body dynamics model with a neural network compensator for friction and other unmodeled effects. Wu \cite{wu2012semi} introduced a semi-parametric \ac{gp} framework for robot dynamics identification. Although these methods improve model accuracy, they face practical challenges: neural-network and \ac{gp} parameterizations are typically high-dimensional and non-convex; and translating their learned residual models and uncertainty into robust contact-force estimates often requires additional assumptions.

Recent advances incorporate uncertainty quantification within \ac{akf} frameworks. Wei et al.~\cite{wei2023contact} employed sparse \acp{gp} to jointly estimate the mean and uncertainty of residual torques, integrating these into an \ac{adkf} observer---an \ac{akf} specialization for disturbance estimation. While this adapts the Kalman gain from data-driven uncertainty, it captures only epistemic uncertainty (due to data sparsity) and an assumed observation noise, ignoring the state-dependent (heteroscedastic) variance that strongly affects the estimates  given the highly nonlinear residual torques. Moreover, sparse-\ac{gp}-based methods, though computationally efficient relative to full \acp{gp}, may still fall short of real-time requirements in high-dimensional or high-frequency robot control. Kim et al.~\cite{kim2026task} proposed a Bayesian-augmented interacting-multiple-model disturbance Kalman filter for contact force estimation, where a Bayesian neural network compensates residual dynamics and its predictive variance serves as state-dependent process noise to handle varying external force dynamics.

We therefore adopt \ac{kmp} to model residual torques. Unlike \ac{gp} predictive variance, which serves primarily as a model-confidence proxy under common noise assumptions, \ac{kmp} provides a state-dependent predictive variance that we use as a heteroscedastic confidence term in the filter, without decomposing it into epistemic and aleatoric components. \ac{kmp} has shown success in imitation learning~\cite{silverio2019uncertainty, winter2024state}, human-robot collaboration~\cite{qian2025hierarchical}, and obstacle avoidance~\cite{xiao2024kmp}, yet to the best of our knowledge, its simultaneous variance modeling and adaptability have not been utilized for system identification or force observer design. This dual uncertainty coverage makes \ac{kmp} a compelling foundation for robust and reliable sensorless force estimation.

Beyond robot dynamics modeling, observer design for force and contact estimation continues to advance. Strategies include momentum observers and \acp{dob}~\cite{zhu2018adaptive}, their nonlinear variant \acp{ndo}~\cite{chen2000nonlinear}, and \acp{gmo}~\cite{vorndamme2017collision}. Ren et al.~\cite{ren2026sensorless} recently introduced a physical–neural adaptive super-twisting momentum observer that combines a rigid-body dynamic model with neural-network residual compensation and a super-twisting sliding-mode structure for robustness under modeling uncertainties. For stochastic estimation and multi-sensor fusion, \ac{ekf} was used to form high-dimensional virtual sensors for external wrench estimation~\cite{roveda20206d}, while \ac{dkf}~\cite{hu2017contact} and neural-network–augmented variants~\cite{liu2021sensorless} blend model-based and data-driven elements. Although \ac{dob} and \ac{ndo} perform well, their effectiveness is inherently tied to the structural disturbance assumptions they impose~\cite{wei2023contact, liu2021sensorless, kim2026task, ren2026sensorless}. Consequently, recent work has shifted toward residual-learning and adaptive filtering frameworks that explicitly model and compensate state-dependent uncertainties. The proposed \ac{kvark} follows and extends this paradigm by incorporating heteroscedastic variance modeling within the adaptive disturbance filtering process. Table~\ref{tab:comparison} summarizes the key differences between our proposed method \ac{kvark} and representative literature approaches. Combining \ac{kmp} with an \ac{adkf}, our method enables a unified treatment of both epistemic and aleatoric uncertainty, with real-time computational practicality and interpretability for safety-critical applications.

\begin{table*}[t]
\centering
\renewcommand{\arraystretch}{1.0}
\setlength{\tabcolsep}{3pt}
\footnotesize
\caption{Notation Conventions Used Throughout the Paper}
\label{tab:notations}
\begin{tabular}{@{}c c p{3.9cm} c c p{3.9cm} c c p{4.1cm}@{}}
\toprule
\textbf{Symbol} & & \textbf{Meaning} & \textbf{Symbol} & & \textbf{Meaning} & \textbf{Symbol} & & \textbf{Meaning} \\
\midrule
$\bm{\omega}$ & $\defeq$ & State vector & $\bm{S}$ & $\defeq$ & State transition matrix & $\bm{w}$ & $\defeq$ & Process noise vector \\
$\bm{\Sigma}$ & $\defeq$ & Noise covariance matrix & $\bm{y}$ & $\defeq$ & Measurement vector & $\bm{H}$ & $\defeq$ & Measurement matrix \\
$\bm{\nu}$ & $\defeq$ & Measurement noise vector & $\bm{P}$ & $\defeq$ & Error covariance matrix & $\bm{K}$ & $\defeq$ & Kalman gain \\
$\rho$ & $\defeq$ & Forgetting factor & $\bm{r}$ & $\defeq$ & Adaptation rate & $m$ & $\defeq$ & Number of measurements \\
\midrule
$\{ \cdot\}_\mathrm{m}$ & $\defeq$ & Subscript for motor/commanded & $\{ \cdot\}_\mathrm{ext}$ & $\defeq$ & Subscript for external effects & $\{ \cdot\}_\mathrm{f}$ & $\defeq$ & Subscript for friction \\
$\{ \cdot\}_\mathrm{dist}$ & $\defeq$ & Subscript for disturbance & $\{ \cdot\}_\mathrm{r}$ & $\defeq$ & Subscript for residual & $\{ \cdot\}_\mathrm{d}$ & $\defeq$ & Subscript for process model \\
$\{ \cdot\}_\nu$ & $\defeq$ & Subscript for measurement model & $\{ \cdot\}_\mathrm{EL}$ & $\defeq$ & Subscript for nominal & $\{ \cdot\}_\mathrm{emp}$ & $\defeq$ & Subscript for empirical \\
$\{ \cdot\}_*$ & $\defeq$ & Query Point & $\hat{\{ \cdot\}}$ & $\defeq$ & Estimated version/value & $\bar{\{ \cdot\}}$ & $\defeq$ & Midpoint \\
\midrule
$\bm{s}$ & $\defeq$ & Input vector of the demonstrations & $\bm{\xi}$ & $\defeq$ & Output vector of demonstrations & $\{ \cdot\}^{(v)}$ & $\defeq$ & Demonstration index $v=1,\dots,V$ \\
$\{ \cdot\}^{(i)}$ & $\defeq$ & Trajectory distribution index $i=1,\dots,N$ & $\{ \cdot\}^d$ & $\defeq$ & Dimension of input vector of demonstrations & $\{ \cdot\}^o$ & $\defeq$ & Dimension of output vector of demonstrations \\
\midrule
$\mathscr{K}$ & $\defeq$ & Kernel gram matrix & $\bm{k}$ & $\defeq$ & Kernel vector & $l$ & $\defeq$ & Kernel length scale \\
$\lambda_1,\lambda_2$ & $\defeq$ & KMP regularization parameters & & & & & & \\
\midrule
$\mathcal{IW}$ & $\defeq$ & Inverse Wishart distribution & $\bm{\Upsilon}$ & $\defeq$ & Scale matrix & $\bm{\mu}$ & $\defeq$ & Mean \\
$\lambda$ & $\defeq$ & Degree of freedom parameter & & & & & & \\
\midrule
$\{ \cdot\}_k$ & $\defeq$ & Time step index & $t_\mathrm{s}$ & $\defeq$ & Sampling period & $f_\mathrm{s}$ & $\defeq$ & Sampling rate \\
$t$ & $\defeq$ & Time & $\zeta$ & $\defeq$ & External torque (discretized) & $\bm{e}$ & $\defeq$ & State transition error \\
\bottomrule
\end{tabular}
\end{table*}

\section{Background}\label{sec:background}
Throughout the paper, vectors are denoted in bold lowercase, matrices in bold uppercase, scalars in italic, and distributions in calligraphic style. Subscripts and superscripts indicate indices, labels, dimensions, and special cases (e.g., estimates, virtual quantities, or query points). A summary of these conventions is provided in Table~\ref{tab:notations}.

\subsection{Robot Manipulator Dynamics}

In this paper, the robot is modeled as an open-chain rigid-body system with $n$ revolute joints, characterized by the configuration vector $\bm{q} \in \mathbb{R}^n$ and its derivatives $\dot{\bm{q}}, \ddot{\bm{q}}$ denoting joint velocities and accelerations, respectively. The system’s dynamics are described by the Euler-Lagrange formulation~\cite{spong2008robot}:
\begin{equation}
  \bm{M}(\bm{q})\ddot{\bm{q}} + \bm{C}(\bm{q}, \dot{\bm{q}})\dot{\bm{q}} + \bm{g}(\bm{q}) = \bm{\tau}_\mathrm{m} - \bm{\tau}_\mathrm{ext} - \bm{\tau}_\mathrm{r},
  \label{eq:EL}
\end{equation}
where $\bm{M}(\bm{q})$ is the symmetric, positive-definite inertia matrix; $\bm{C}(\bm{q}, \dot{\bm{q}})$ incorporates Coriolis and centrifugal effects; $\bm{g}(\bm{q})$ is the gravitational torque vector; $\bm{\tau}_\mathrm{m}$ denotes the commanded motor torque; $\bm{\tau}_\mathrm{ext}$ represents the external torque; and $\bm{\tau}_\mathrm{r}$ captures unmodeled effects, collectively referred to as the residual torque.

The residual torque $\bm{\tau}_\mathrm{r}$ combines frictional forces, elasticities, actuator/sensor peculiarities, and other unmodeled \changeOLD{internal} dynamics while external disturbances such as payload-induced effects and external contacts are either modeled inside the nominal dynamics or estimated as in \cite{haddadin2017robot}. Reliable estimation of this term is essential, as it substantially impacts both control fidelity and force estimation. Further, $\bm{\tau}_\mathrm{r}$ is decomposed into physically meaningful contributions:
\begin{equation}
  \bm{\tau}_\mathrm{r} \coloneqq \bm{\tau}_\mathrm{f} + \bm{\tau}_\mathrm{dist},
\end{equation}
where $\bm{\tau}_\mathrm{f}$ captures joint friction and $\bm{\tau}_\mathrm{dist}$ aggregates other disturbances---both are generally state-dependent and stochastic, underscoring the need for probabilistic modeling.

\subsection{Discrete Momentum Model}
Direct numerical differentiation of encoder signals to calculate $\ddot{\bm{q}}$ is highly sensitive to noise; thus, for this work, a momentum-based approach is preferred \cite{de2003actuator}. Here, the system’s energy-shaping properties are made explicit:

\noindent\textit{Skew-symmetry Property.}  
A central property of robot dynamics is the skew-symmetry of $\dot{\bm{M}}(\bm{q}) - 2\bm{C}(\bm{q},\dot{\bm{q}})$ for all $\bm{q}, \dot{\bm{q}}$~\cite{spong2008robot}. That is, for any $\bm{z}\in\mathbb{R}^{n}$:
\begin{equation}
\bm{z}^{\top}\left(\dot{\bm{M}}(\bm{q}) - 2\bm{C}(\bm{q},\dot{\bm{q}})\right)\bm{z} = 0.
\end{equation}
This constraint ensures energy conservation, so Coriolis and centrifugal terms only redistribute kinetic energy and do not inject or dissipate it. The kinetic energy of the manipulator is defined as:
\begin{equation}
k = \frac{1}{2}\dot{\bm{q}}^{\top}\bm{M}(\bm{q})\dot{\bm{q}}.
\end{equation}
External joint torques are solely responsible for changes in system energy.
Introducing the generalized momentum:
\begin{equation}
\bm{p} = \bm{M}(\bm{q})\dot{\bm{q}},
\end{equation}
and differentiating, the system dynamics are recast as:
\begin{equation}
  \dot{\bm{p}} = \bm{C}^{\top}(\bm{q},\dot{\bm{q}})\dot{\bm{q}} - \bm{g}(\bm{q}) + \bm{\tau}_\mathrm{m} - \bm{\tau}_\mathrm{ext} - \bm{\tau}_\mathrm{r}.
\end{equation}

This reformulation avoids amplification of noise from acceleration estimates, \changeOLD{thereby increasing the robustness of the estimation.}

A compact state-space description is then:
\begin{equation}
  \dot{\bm{x}} = \underbrace{\bm{0}}_{\bm{A}}\bm{x} + \underbrace{\bm{I}}_{\bm{B}}\bm{u} + \bm{D}_1\bm{d}_1 + \bm{D}_2\bm{d}_2,
\end{equation}
with:
\begin{itemize}
  \item $\bm{x} \coloneqq \bm{p}$,
  \item $\bm{u} \coloneqq \bm{C}^{\top}(\bm{q},\dot{\bm{q}})\dot{\bm{q}} - \bm{g}(\bm{q}) + \bm{\tau}_\mathrm{m}$,
  \item $\bm{d}_1 \coloneqq \bm{\tau}_\mathrm{ext}$, $\bm{d}_2 \coloneqq \bm{\tau}_\mathrm{r}$,
  \item $\bm{D}_1 = \bm{D}_2 = -\bm{I}$.
\end{itemize}
Here $\bm{A}, \bm{B},\bm{D}$ are system matrices and $\bm{u},\bm{d}$ are input vectors.
Discretizing with sampling period $t_\mathrm{s}$ via forward Euler, the model becomes:
\begin{equation}
  \bm{x}_k = \bm{x}_{k-1} + t_\mathrm{s}\bm{u}_{k-1} - t_\mathrm{s}\bm{\tau}_{\mathrm{ext},k-1} - t_\mathrm{s}\bm{\tau}_{\mathrm{r},k-1}.
  \label{eq:state_dt}
\end{equation}
This equation underpins subsequent \ac{kf} formulations for sensorless force estimation.

\subsection{Kalman Filter Framework}
\label{subsec:KF}
The standard \ac{kf} is a recursive Bayesian estimator for linear systems with Gaussian noise, optimal for propagating both state estimates and the associated uncertainty. These filters underpin most modern state estimators and force observers in robotics due to their balance of computational efficiency and statistical rigor.

At each time step $k$, the KF operates in two stages:
\begin{itemize}
    \item {Prediction:} Forward propagate the current state estimate using the system model.
    \item {Update:} Incorporate new sensor data, correcting the estimate via the measurement innovation.
\end{itemize}

\noindent\textit{Process model.}
The general form of the linear process (state transition) model is:
\begin{equation}
\bm{\omega}_k = \bm{S} \bm{\omega}_{k-1} + \bm{w}_{\mathrm{d},k-1},
\end{equation}
where $\bm{\omega}_k \in \mathbb{R}^n$ is the state vector, $\bm{S} \in \mathbb{R}^{n \times n}$ is the state transition matrix, and $\bm{w}_{\mathrm{d},k-1} \sim \mathcal{N}(\bm{0}, \bm{\Sigma}_{\mathrm{d},k-1})$ is the zero-mean Gaussian noise with covariance $\bm{\Sigma}_{\mathrm{d},k-1}$.

\noindent\textit{Measurement model.}
Measurements $\bm{y}_k$ are related to the state through:
\begin{equation}
\bm{y}_k = \bm{H} \bm{\omega}_{k} + \bm{\nu}_k,
\end{equation}
where $\bm{H} \in \mathbb{R}^{m \times n}$ is the measurement matrix, and $\bm{\nu}_k \sim \mathcal{N}(\bm{0}, \bm{\Sigma}_{\mathrm{\nu},k})$ is Gaussian measurement noise with covariance $\bm{\Sigma}_{\mathrm{\nu},k}$.

\noindent\textit{Kalman filter recursion.}
The standard \ac{kf} recursions are:
\begin{equation}
\begin{aligned}
	\hat{\bm{\omega}}_{k|k-1} &= \bm{S} \hat{\bm{\omega}}_{k-1|k-1}, \\
	\bm{P}_{k|k-1} &= \bm{S} \bm{P}_{k-1|k-1} \bm{S}^\top + \bm{\Sigma}_{\mathrm{d},k-1}, \\
	\bm{K}_k &= \bm{P}_{k|k-1} \bm{H}^\top \left( \bm{H} \bm{P}_{k|k-1} \bm{H}^\top + \bm{\Sigma}_{\mathrm{\nu},k} \right)^{-1}, \\
	\hat{\bm{\omega}}_{k|k} &= \hat{\bm{\omega}}_{k|k-1} + \bm{K}_k \left( \bm{y}_k - \bm{H} \hat{\bm{\omega}}_{k|k-1} \right), \\
	\bm{P}_{k|k} &= \left( \bm{I} - \bm{K}_k \bm{H} \right) \bm{P}_{k|k-1}.
\end{aligned}
\label{eq:standardKF}
\end{equation}

\noindent\textit{Adaptive Kalman filter.}
In practice, the process and measurement covariances $\bm{\Sigma}_{\mathrm{d},k}, \bm{\Sigma}_{\mathrm{\nu},k}$ are unknown or time-varying. \ac{akf} schemes \cite{khodarahmi2023review, rutan1991adaptive} adapt these covariances online, often leveraging innovation-based updates such as \cite{mohamed1999adaptive}:
\begin{equation}
\label{eq:akf}
\begin{split}
	\bm{\Sigma}_{\mathrm{\nu},k}
	&= (1 - \rho_\mathrm{\nu}) \bm{\Sigma}_{\mathrm{\nu},{k-1}} \\
	&\quad + \rho_\mathrm{\nu}
	\left(
	\bm{r}_k \bm{r}_k^\top
	+
	\bm{H} \bm{P}_{k|k-1} \bm{H}^\top
	\right),
\end{split}
\end{equation}
and
\begin{equation}
	\bm{\Sigma}_{\mathrm{d},k} = (1 - \rho_\mathrm{d}) \bm{\Sigma}_{\mathrm{d},k-1} + \rho_\mathrm{d} \bm{K}_k \bm{r}_k \bm{r}_k^\top \bm{K}_k^\top,
\end{equation}

where ${\rho_\mathrm{\nu}, \rho_\mathrm{d} \in (0,1]}$ are forgetting factors that control the adaptation rate and $\bm{r}_k = \bm{y}_k - \bm{H} \hat{\bm{\omega}}_{k|k-1}$ is the innovation.

This adaptive filtering framework is critical for robust external force estimation in the presence of model uncertainties. Later sections instantiate these models with robot-specific dynamics and virtual measurements, leveraging \ac{kmp}-derived mean and uncertainty.

\subsection{\changeOLD{Kernelized} Movement Primitives}
\label{subsec:KMP}

\ac{kmp} \cite{huang2019kernelized} provides a nonparametric probabilistic framework for learning and generalizing trajectory distributions from demonstrations, leveraging kernel methods to model input–output correspondences in high-dimensional spaces without the constraints of fixed basis functions.

Given a set of demonstration pairs $\{ (\bm{s}^{(v)}, \bm{\xi}^{(v)}) \}_{v=1}^V$, where $\bm{s}^{(v)} \in \mathbb{R}^d$ (e.g., time, joint state, or velocity in $d$ dimensions) and $\bm{\xi}^{(v)} \in \mathbb{R}^o$ (e.g., joint position, velocity, or torque in $o$ dimensions), KMP encodes each demonstration probabilistically as an $N$-dimensional probabilistic trajectory distribution $\{\bm{s}^{(i)},\hat{\bm{\mu}}^{(i)},\hat{\bm{\Sigma}}^{(i)} \}_{i=1}^N$ typically extracted from a \ac{gmr} on data modeled by a \ac{gmm}. This yields a reference database—a continuous probabilistic representation of the demonstrated trajectories.

For prediction, \ac{kmp} constructs a block-diagonal covariance matrix and a stacked mean vector:
\begin{equation} 
	\label{eq:KMP_14}
	\begin{split}
	    \bm{\Sigma} =& \operatorname{blockdiag}(\hat{\bm{\Sigma}}^{(1)}, \ldots,\hat{\bm{\Sigma}}^{(i)}, \ldots, \hat{\bm{\Sigma}}^{(N)}), 
	\quad \\
	\bm{\mu} =& 
	\begin{bmatrix}
        \hat{\bm{\mu}}^{(1)\top} & \cdots &
		\hat{\bm{\mu}}^{(i)\top} & \cdots & \hat{\bm{\mu}}^{(N)\top}
	\end{bmatrix}^\top ,
	\end{split}
\end{equation}
where $\bm{\Sigma} \in \mathbb{R}^{oN \times oN}$ and $\bm{\mu} \in \mathbb{R}^{oN}$.

Then, the kernel Gram matrix $\mathscr{K} \in \mathbb{R}^{oN \times oN}$ 
and the query–to–training kernel vector $\bm{k}_* \in \mathbb{R}^{oN \times o}$ are defined:
\begin{equation}
	\label{eq:KMP_19}
	\mathscr{K}_{ij} = \bm{k}(\bm{s}^{(i)}, \bm{s}^{(j)}), 
	\quad i,j = 1,\dots,N,
\end{equation}
\begin{equation}
	\label{eq:KMP_20}
	\bm{k}_* = 
	\begin{bmatrix}
		\bm{k}(\bm{s}_*, \bm{s}^{(1)}) & \cdots & \bm{k}(\bm{s}_*, \bm{s}^{(N)})
	\end{bmatrix}^\top ,
\end{equation}
where $ \bm{k}(\bm{s}^{(i)}, \bm{s}^{(j)}) = k(\bm{s}^{(i)}, \bm{s}^{(j)})\bm{I}_o$ and $k(\cdot,\cdot)$ is the chosen positive–definite kernel function. Here we use squared exponential kernel:
\begin{equation}
	k(\bm{s}^{(i)}, \bm{s}^{(j)}) 
	= \sigma_f^2 \exp\left( 
	-\frac{1}{2} (\bm{s}^{(i)} -  \bm{s}^{(j)})^\top 
	l^{-1} (\bm{s}^{(i)} -  \bm{s}^{(j)}) 
	\right)
\end{equation}
which is a common choice in literature. \ac{kmp} derives the predictive mean and covariance by minimizing the \ac{kl} divergence between a parameterized trajectory distribution and the probabilistic reference trajectory. The predictive mean for a query input $\bm{s}_*$ is computed as:
\begin{equation}\label{eq:KMP_mean}
	\mathbb{E}[\bm{\xi}(\bm{s}_*)] 
	=\bm{\mu}_*= \bm{k}_*^\top (\mathscr{K} + \lambda_1 \bm{\Sigma})^{-1} \bm{\mu},
\end{equation}
and the predictive covariance as:
\begin{equation}\label{eq:KMP_var}
	\operatorname{Cov}[\bm{\xi}(\bm{s}_*)] 
	=\bm{\Sigma}_*= \frac{N}{\lambda_2} 
	\left( \bm{k}_{**} - \bm{k}_* (\mathscr{K} + \lambda_2 \bm{\Sigma})^{-1} \bm{k}_* \right),
\end{equation}
where $\bm{k}_{**} = \bm{k}(\bm{s}_*, \bm{s}_*)$, and $\lambda_1, \lambda_2 > 0$ are regularization terms controlling mean fitting and variance scaling, respectively. Hence, \ac{kmp} requires the definition of 4 hyperparameters $\{\sigma_f^2,l,\lambda_1,\lambda_2\}$. Due to its kernel formulation, \ac{kmp} naturally handles high–dimensional inputs and enables extrapolation beyond training regions. Furthermore, this kernel-based construction empowers KMP to directly address both epistemic (data scarcity) and aleatoric (data variability) uncertainty, yielding smooth interpolation within the demonstrated region and principled uncertainty extrapolation outside it. 

\section{Proposed Method}\label{sec:proposed method}
This section presents the \ac{kvark} framework for sensorless external force estimation in robotic manipulators. \ac{kvark} first establishes data-driven models of joint residual torques by leveraging \ac{kmp} to learn the mean and variance of residual dynamics from richly excited trajectories free of external disturbances such as payloads or external forces. These probabilistic estimates serve as adaptive virtual measurements within an \ac{adkf} that estimates external joint torques while accounting for unknown, time-varying process noise from unmodeled sources. \ac{kvark} is modular, with each component addressing a distinct limitation: residual-torque learning reduces systematic modeling bias, \ac{kmp} provides a probabilistic (uncertainty-aware) virtual measurement, and \ac{vb}-based process-noise adaptation improves robustness under contact transitions and time-varying disturbances. 

\subsection{Data Acquisition and Feature Selection}

The initial step in our method is to use the \ac{kmp} framework to approximate the joint-wise residual torques. To ensure the accuracy of our residual torque model, we require a dataset that thoroughly and uniformly spans the robot's dynamic state space, $\mathbb{Z} = \{(\bm{q}, \dot{\bm{q}}, \ddot{\bm{q}})\}$, while respecting the joint position, velocity, and acceleration limits.

To achieve this, we excite the robot with a series of optimized random-Fourier trajectories, as described in \cite{dong2021dynamic}:
\begin{equation} \label{eq:fourier_trajectory}
    \bm{q}^{(v)}_j(t)
    = \bar{q}_j
    + \sum_{\kappa=1}^{K}
      \bigl[
        a^{(v)}_{j,\kappa}\,\sin(\omega_{\kappa} t)
        + b^{(v)}_{j,\kappa}\,\cos(\omega_{\kappa} t)
      \bigr],
\end{equation}
where $v$ and $j$ are the indices for the trajectory instance and the joint, respectively, $K$ is the number of harmonics, $\omega_{\kappa}$ is the $\kappa$-th harmonic frequency over the trajectory duration, $\bar{q}_j$ is the midpoint of the joint's range of motion, and the coefficients $(\bm{a}, \bm{b})$ form the parameter vector $\bm{\theta} \in \mathbb{R}^{2nK}$.

To ensure comprehensive yet safely executable state-space coverage, we define the state cloud generated by $\bm{\theta}$ as $\mathcal{S}(\bm{\theta}) = \{\bm{z}_t(\bm{\theta})\}_{t=1}^{T} \subset \mathbb{R}^{3n}$. The fixed trajectory parameters in \eqref{eq:fourier_trajectory} are selected to excite the low- and moderate-velocity regimes where friction and residual torque effects are most pronounced, while avoiding excessive accelerations and remaining within the controller bandwidth. We then employ the log-determinant of the sample covariance as our objective function:
\begin{equation}
  J(\bm{\theta})
  =-\log\det\bigl(\operatorname{Cov}(\mathcal{S}(\bm{\theta}))\bigr).
\end{equation}
Minimizing $J(\bm{\theta})$ maximizes the volumetric dispersion of the state cloud. The optimization is constrained by joint limits and collision-avoidance inequalities and solved using a \ac{ga}. The numerical trajectory parameters and \ac{ga} settings used in the experiments are reported in Section~\ref{sec:experiments}.

\noindent\textit{Execution and recording.} 
Each trajectory is executed in a contact-free environment under low-level position control. At a sampling rate of $f_\mathrm{s} = 1/t_\mathrm{s}$, we record the following:
\begin{equation}
      \bigl\{\{
    \bm{q}_{t}^{(v)},\,
    \dot{\bm{q}}_{t}^{(v)},\,
    \ddot{\bm{q}}_{t}^{(v)},\,
    \bm{\tau}_{\mathrm{m},t}^{(v)}\}_{t=1}^{T}
  \bigr\}_{v=1}^{V}.
\end{equation}

The residual torques are then computed online using:
\begin{equation}
	\label{eq:residual_torque}
	\bm{\tau}_{\mathrm{r},t}
	= \bm{\tau}_{\mathrm{m},t}
    - \bm{\tau}_{\mathrm{EL},t},
\end{equation}
where $\bm{\tau}_{\mathrm{EL}}$ represents the nominal torques computed from a preloaded rigid-body model initialized from the parameters extracted from the manufacturer-provided URDF as:
\begin{equation*}
	\bm{M}(\bm{q})\ddot{\bm{q}}
    + \bm{C}(\bm{q}, \dot{\bm{q}})\dot{\bm{q}}
    + \bm{g}(\bm{q})
    = \bm{\tau}_{\mathrm{EL}}.
\end{equation*}

These residual torques, paired with their corresponding inputs, are then passed to the \ac{kmp} pipeline. Specifically, for each time step and joint we form input–output pairs $\bm{s}_{t,j} \coloneqq (\bm{q}_{t,j}, \dot{\bm{q}}_{t,j}, \ddot{\bm{q}}_{t,j})$ and $\bm{\xi}_{t,j} \coloneqq \bm{\tau}_{\mathrm{r},t,j}$, which we subsequently reindex as demonstrations  $\{ (\bm{s}^{(v)}, \bm{\xi}^{(v)}) \}_{v=1}^V$. The optimized data collection ensures that the dataset spans the state space $\mathbb{Z}$ almost isotropically, which greatly enhances the generalization capabilities of the learned residual model.

\begin{remark}
While our initial approach utilized a high-dimensional input space (joint positions, velocities, and accelerations) for modeling the residual torques, this led to poor generalization due to overfitting in sparse regions of the feature space. To mitigate this, we have restricted the input to only the joint velocity. This not only reflects the physical dependence of friction on motion but also allows the KMP model to generalize more effectively while maintaining high modeling fidelity. A detailed analysis, including empirical comparisons and \ac{ard}-based feature relevance, is provided in the Discussion section.
\end{remark} 

\subsection{Residual Torque Modeling with \ac{kmp}}

With the considerations mentioned above, we model the joint-wise residual torque, ${\tau}_{\mathrm{r},j}$, as a function of the joint velocity, ${\dot{q}}_j$, using \ac{kmp}. Each joint is treated independently, and a one-dimensional input space is used for training the \ac{kmp} model. The model is initialized through \ac{gmr} performed on the recorded dataset.

The resulting probabilistic reference trajectory provides both mean and variance profiles at selected support points. \ac{kmp} leverages these to generate smooth and consistent estimates of the residual torques. At test time, \ac{kmp} provides two key outputs: the predicted mean of the residual torque and its associated variance \eqref{eq:KMP_mean}-\eqref{eq:KMP_var}. The mean is used to correct the dynamics model, while the variance provides a unified quantification of modeling uncertainty and variability.

This is made possible by the kernel-based structure of \ac{kmp}. The predicted variance naturally increases as the test input, $\bm{s}_*$, moves away from the training data. This behavior can be understood by examining the properties of the squared exponential kernel, $k(\cdot, \cdot)$, used in the \ac{kmp} model. For a test input  $\bm{s}_*$ and any training input $\bm{{s}}^{(i)}$, the kernel value satisfies:
\[
\lim_{\|\bm{s}_* - \bm{{s}}^{(i)}\| \to \infty} k(\bm{s}_*, \bm{{s}}^{(i)}) = 0.
\]
This implies that the kernel vector, $\bm{k}_*$, becomes vanishingly small far from the training points. As a result, the predictive variance converges to a constant matrix determined by the kernel hyperparameters and the reference trajectory's structure:
\[
\lim_{\|\bm{s}_* - \bm{{s}}^{(i)}\| \to \infty} \operatorname{Cov}[\bm{\xi}(\bm{s}_*)] = \frac{N}{\lambda_2} \sigma_f^2,
\]
where $\sigma_f^2$ is the kernel amplitude and $\lambda_2$ is the variance regularization term.

This asymptotic behavior is central to distinguishing between data-driven variability (aleatoric uncertainty) and lack of knowledge (epistemic uncertainty). Within the training region, the predicted variance reflects the heteroscedastic noise present in the data. Outside the training region, however, the variance grows to signal a lack of knowledge.

This dual nature of the variance is particularly advantageous for residual torque modeling. In regions with dense training data, the variance characterizes the repeatability of the residual torque, providing insight into dynamic variability or sensor noise. Conversely, in regions with sparse or no training data, the variance increases systematically due to kernel decay, indicating higher epistemic uncertainty.

By leveraging the predictive structure of \ac{kmp}, we obtain a model that not only estimates the expected residual torques but also quantifies their reliability. This variance information can then be propagated to downstream state estimators, such as our adaptive Kalman filter, to improve robustness and safety.

The uncertainty in \ac{kmp} can be finely controlled through the hyperparameters $\sigma_f^2$, $N$, and $\lambda_2$. Increasing $\sigma_f^2$ or $N$ scales the asymptotic uncertainty, making the model more conservative in data-sparse regions. In contrast, increasing $\lambda_2$ suppresses uncertainty growth, making the model more confident during extrapolation. These hyperparameters do not directly affect the predictive mean. This decoupling lets \ac{kmp}'s uncertainty behavior be tuned independently of its fit quality. While other kernelized methods (e.g., \ac{gpr}) yield uncertainty estimates, they lack this decoupling. In \ac{gpr}, the predictive variance converges to the kernel variance $\sigma_f^2$, but $\sigma_f^2$  directly influences the mean prediction. In contrast, the predictive mean in \ac{kmp} depends only on kernel similarities and the \ac{gmr}-derived reference means, while the uncertainty is modulated by $\lambda_2$ and the reference distribution's structure. This separation makes \ac{kmp} more robust to hyperparameter tuning and better suited for applications demanding accurate modeling and well-calibrated uncertainty.

\subsection{\ac{kvark}  Formulation}

Having established a probabilistic model of residual torques using \ac{kmp}, we integrate these predictions into a state estimator for external force reconstruction. We formulate an observer that leverages the mean and variance outputs from \ac{kmp}, resulting in a kernel-augmented \ac{akf}---termed \ac{kvark}.

\noindent\textit{Problem Setup.} 
Starting from the discrete-time momentum-based dynamics in \eqref{eq:state_dt}, the residual torque is now modeled by the predictive mean from the \ac{kmp}, denoted as $\bm{\mu}_*$. We rearrange the dynamics to isolate the external torque:
\begin{equation}
\label{eq:virtual_measurement}
    {\bm{\zeta}}^*_k \coloneqq \bm{x}_k - \bm{x}_{k-1} - t_\mathrm{s} \bm{u}_{k-1} + t_\mathrm{s} \bm{\mu}_{*,k} \approx -t_\mathrm{s} \bm{\tau}_{\mathrm{ext},k-1}.
\end{equation}
This quantity, $ {\bm{\zeta}}^*_k $, is a \emph{virtual measurement} of the external torque compensated for modeled residual friction effects. This measurement inherits uncertainty originating from sensor noise and the predictive variance of the \ac{kmp} model. Our measurement model then becomes:
\begin{equation}
    {\bm{\zeta}}^*_k = \bm{H} \bm{\omega}_{k} + \bm{\nu}_k, \qquad \bm{H} = -t_\mathrm{s} \bm{I}_n,
\end{equation}
where $\bm{\omega}_{k} \coloneqq \bm{\tau}_{\mathrm{ext},k}$ is the state to be estimated and $\bm{\nu}_k$ is a zero-mean Gaussian noise term with covariance $\bm{\Sigma}_{\mathrm{\nu},k}$.

\noindent\textit{Measurement Noise Modeling.}
To account for uncertainty in residual torque estimation, we define the total measurement noise covariance as the sum of two components:
\begin{equation}
\label{eq:noise_cov}
    \bm{\Sigma}_{\mathrm{\nu},k} = t_\mathrm{s}^2 \bm{\Sigma}_{*,k} + \bm{\Sigma}_{\mathrm{emp},k},
\end{equation}
where $\bm{\Sigma}_{*,k}$ is the predictive covariance of \ac{kmp} at the current input, and $\bm{\Sigma}_{\mathrm{emp},k}$ represents the empirical noise.

The \ac{kmp}-derived term, $\bm{\Sigma}_{*,k}$, captures both heteroscedasticity in the training data and epistemic uncertainty, as discussed in Section~\ref{subsec:KMP}. While \ac{kmp} provides an offline, structured characterization of variability, it cannot adapt to time-varying measurement noise, numerical-differentiation artifacts, or unexpected disturbances during online operation. To address this limitation, the empirical noise covariance $\bm{\Sigma}_{\mathrm{emp},k}$ is adapted online from filter innovations using an exponentially weighted moving average, which is a specialized form of~\eqref{eq:akf}:
\begin{equation}
\label{eq:noise_adaptation}
    \bm{\Sigma}_{\mathrm{emp},k} = \bm{\Sigma}_{\mathrm{emp},k-1} + \rho \left( \operatorname{diag}(\bm{r}_k^2) - \bm{\Sigma}_{\mathrm{emp},k-1} \right),
\end{equation}
where $\bm{r}_k = \bm{\zeta}^*_k - \bm{H} \hat{\bm{\omega}}_{k|k-1}$ denotes the innovation and $\rho \in (0,1)$ is the forgetting factor.

\begin{remark}
Although this online adaptation improves robustness, the KMP variance $\bm{\Sigma}_{*,k}$ already absorbs a portion of the measurement noise present in the training data. Naively summing $\bm{\Sigma}_{*,k}$ and $\bm{\Sigma}_{\mathrm{emp},k}$ would therefore count the stationary noise component twice. To mitigate this, $\bm{\Sigma}_{\mathrm{emp},k}$ is initialized and upper-bounded such that it cannot exceed the median of $\bm{\Sigma}_{*,k}$ on the training set, ensuring the empirical term refines rather than dominates the state-dependent KMP variance.
\end{remark}

\noindent\textit{Process Model.}
We model the external torque dynamics, $\bm{\omega}_k$, as a first-order random walk to allow flexibility in tracking unknown or time-varying external forces:
\begin{equation}
    \bm{\omega}_k = \bm{\omega}_{k-1} + \bm{w}_{\mathrm{d},k-1}, \quad \bm{w}_{\mathrm{d},k-1} \sim \mathcal{N}(\bm{0}, \bm{\Sigma}_{\mathrm{d},k-1}),
    \label{eq:randomWalk}
\end{equation}

This formulation leads to a \ac{kf} where both the process and measurement models are probabilistic, and the measurement uncertainty explicitly incorporates the confidence level of the \ac{kmp} model. Unlike traditional force observers, our approach dynamically adapts its trust in virtual measurement based on epistemic confidence in the residual torque estimation.

The predictive variance of \ac{kmp}, $\bm{\Sigma}_{*,k}$, acts as a gating signal for the measurement's reliability. When the joint velocity lies within a well-supported region of the training set, $\bm{\Sigma}_{*,k}$ is low, and the measurement $\bm{\zeta}^*_k$ is trusted. In contrast, in extrapolated regions or under uncertain dynamics, $\bm{\Sigma}_{*,k}$ inflates the measurement noise, causing the filter to rely more on the prior. This behavior significantly improves the estimator's robustness under model mismatch or external disturbances.

\subsection{Variational-Bayes Adaptation of Process Covariance}

In the proposed observer, the covariance of the noise in the process $\bm{\Sigma}_{\mathrm{d},k}$ associated with the dynamics of the external torque is not constant but varies according to changes in the behavior of the unmodeled disturbance. To ensure robust tracking and avoid ad hoc heuristic gain tuning, we adopt a \ac{vb}  approach that treats $\bm{\Sigma}_{\mathrm{d},k}$ as a latent random variable and updates its entire distribution online from the filter residuals as described in~\cite{sarkka2009recursive}. Specifically, we place an inverse-Wishart ($\mathcal{IW}$) prior on the covariance matrix $\bm{\Sigma}_{\mathrm{d},k}$:
\begin{equation}
p(\bm{\Sigma}_{\mathrm{d},k}) = \mathcal{IW}(\lambda_{k|k-1}, \bm{\Upsilon}_{k|k-1}),
\end{equation}
where $\lambda_{k|k-1} > n+1$ denotes the prior degrees of freedom, $\bm{\Upsilon}_{k|k-1} \in \mathbb{R}^{n \times n}$ is the scale matrix, and $n$ is the state dimension. The $\mathcal{IW}$ is conjugate to the Gaussian distribution, making it a natural fit for adaptive Bayesian covariance estimation in linear-state-space models~\cite{o2004kendall,koch2008bayesian}.

The total measurement covariance $\bm{\Sigma}_{\mathrm{\nu},k}$, defined in \eqref{eq:noise_cov} as the combination of the \ac{kmp}-predicted residual torque variance $\bm{\Sigma}_{*,k}$ and the empirical sensor noise $\bm{\Sigma}_{\mathrm{emp},k}$, enters the recursive Bayesian update of the process covariance through its influence on the updated covariance $\bm{P}_{k|k}$.

The evolution of the external torque state is modeled as a random walk \eqref{eq:randomWalk} with the likelihood of the observed state transition error $\bm{e}_k$:
\begin{equation}
p(\bm{e}_k \mid \bm{\Sigma}_{\mathrm{d},k}) = \mathcal{N}(\bm{e}_k; \bm{0}, \bm{\Sigma}_{\mathrm{d},k}),
\end{equation}
where the process innovation is $\bm{e}_k = \hat{\bm{\omega}}_{k|k} - \hat{\bm{\omega}}_{k|k-1}$. The term $\hat{\bm{\omega}}_{k|k}$ is obtained using the Kalman-updated precision.
\begin{equation}
\bm{P}_{k|k}^{-1} = \mathbb{E}[\bm{P}_{k|k-1}^{-1}] + \bm{H}^\top \bm{\Sigma}_{\mathrm{\nu},k}^{-1} \bm{H}.
\end{equation}

By applying Bayes’ theorem, the posterior distribution of $\bm{\Sigma}_{\mathrm{d},k}$ after incorporating $\bm{e}_k$ also follows an inverse-Wishart distribution:
\begin{equation}
p(\bm{\Sigma}_{\mathrm{d},k} \mid \bm{e}_k) = \mathcal{IW}(\lambda_{k|k}, \bm{\Upsilon}_{k|k}),
\end{equation}
with the following recursive updates for the parameters:
\begin{equation}
\label{eq:iw_parameter_update}
\begin{aligned}
\lambda_{k|k} &= \lambda_{k|k-1} + 1, \\
\bm{\Upsilon}_{k|k} &= \bm{\Upsilon}_{k|k-1} + \bm{e}_k \bm{e}_k^\top + \bm{P}_{k|k}.
\end{aligned}
\end{equation}

The inclusion of state covariance $\bm{P}_{k|k}$, which is dependent on measurement noise $\bm{\Sigma}_{\mathrm{\nu},k}$, effectively couples the process noise adaptation to the \ac{kmp}-derived measurement uncertainty.

The \ac{vb} estimate of $\bm{\Sigma}_{\mathrm{d},k}$ is then obtained from the posterior mean of the inverse-Wishart distribution:
\begin{equation}
\bm{\Sigma}_{\mathrm{d},k} = \frac{\bm{\Upsilon}_{k|k}}{\lambda_{k|k} - n - 1}.
\label{eq:SigmaIW}
\end{equation}
Substituting \eqref{eq:iw_parameter_update} into \eqref{eq:SigmaIW} yields an explicit recursive form:
\begin{equation}
\label{eq:sigma_d_VB}
\bm{\Sigma}_{\mathrm{d},k} = \frac{\bm{\Upsilon}_{k|k-1} + \bm{e}_k \bm{e}_k^\top + \bm{P}_{k|k}}{\lambda_{k|k-1} + 1 - n - 1}
\end{equation}
This recursion integrates prior knowledge (via $\bm{\Upsilon}_{k|k-1}$ and $\lambda_{k|k-1}$) with new evidence (via $\bm{e}_k \bm{e}_k^\top$ and the measurement-informed $\bm{P}_{k|k}$) in a statistically consistent manner. When $\bm{e}_k$ is large or the measurement uncertainty $\bm{\Sigma}_{\mathrm{\nu},k}$ is high, the scale matrix $\bm{\Upsilon}_{k|k}$ increases dramatically, producing a larger $\bm{\Sigma}_{\mathrm{d},k}$ that allows the filter to respond more aggressively to rapid changes. In contrast, when $\bm{e}_k$ is small and measurements are confident, the covariance contracts, increasing the reliance on the process model and reducing the estimate's variance. Finally, the standard \ac{kf} recursions take the reformulated form:
\begin{equation}
    \begin{aligned}
        \hat{\bm{\omega}}_{k|k-1} &= \bm{I}\,\hat{\bm{\omega}}_{k-1|k-1}\\
        \bm{P}_{k|k-1} &= \bm{I}\,\bm{P}_{k-1|k-1}\,\bm{I}^\top 
        + \bm{\Sigma}_{\mathrm{d},k-1}\\
        \bm{K}_k &= \bm{P}_{k|k-1}(-t_\mathrm{s}\bm{I}_n)^\top \\
        &\quad\times\Big[(-t_\mathrm{s}\bm{I}_n)\,\bm{P}_{k|k-1}(-t_\mathrm{s}\bm{I}_n)^\top 
        + \bm{\Sigma}_{\nu,k}\Big]^{-1}\\
        \hat{\bm{\omega}}_{k|k} &= \hat{\bm{\omega}}_{k|k-1} 
        + \bm{K}_k\big(\bm{\zeta}^*_k 
        - (-t_\mathrm{s}\bm{I}_n)\hat{\bm{\omega}}_{k|k-1}\big)\\
        \bm{P}_{k|k} &= \left[\bm{I}-\bm{K}_k(-t_\mathrm{s}\bm{I}_n)\right]
        \bm{P}_{k|k-1}
    \end{aligned}
    \label{eq:kvark_recursion}
\end{equation}

\begin{algorithm}[t]

\caption{\ac{kvark}}
\label{algorithm1}
\begin{algorithmic}[1]
\Require Training dataset $\mathcal{D} = \{ \dot{\bm{q}}_j, \bm{\tau}_{\mathrm{m},j}\}$; \\
Number of \ac{gmm} components $N$; \\
\ac{kmp} hyperparameters $(\lambda_1, \lambda_2, l, \sigma_f^2)$; \\
Filter parameters: \ac{vb} iterations $M$, forgetting factor $\rho$;\\
Initial covariances $\bm{\Sigma}_{\mathrm{d}},  \bm{\Sigma}_{\mathrm{emp}}$; Sample time $t_\mathrm{s}$
\Ensure Estimated external torque $\{\hat{\bm{\tau}}_{\rm ext,k}\}_{k=1}^\infty$.
\vspace{0.5em}
\State \textbf{\textit{Offline Phase}:} \textit{Train Residual Torque Model}
\For{$j = 1$ to $n$}
    \State Compute residuals: $\bm{\tau}_{\mathrm{r},j} \gets \bm{\tau}_{\mathrm{m},j} - \bm{\tau}_{\mathrm{EL},j}$ using (\ref{eq:residual_torque})
    \State Fit \ac{gmm} to $(\dot{\bm{q}}_j, \bm{\tau}_{\mathrm{r},j})$ data $\rightarrow\{\bm{s}^{(i)},\hat{\bm{\mu}}^{(i)},\hat{\bm{\Sigma}}^{(i)} \}_{i=1}^N$
    \State Train  \ac{kmp} model $\Theta_j$ for joint $j$.
\EndFor
\State \textbf{return} $\bm{\Theta} = \{\Theta_j\}_{j=1}^n$.
\State \textbf{\textit{Online Phase}:} \textit{Real-Time Force Estimation}
\State \textbf{Initialize:} $\hat{\bm{\omega}}_{0|0}\gets \bm{0},\; \bm{P}_{0|0}\gets \bm{I},\; \bm{\Sigma}_{\mathrm{emp},k}\gets\bm{\Sigma}_{\mathrm{emp}}$
\vspace{0.5em}
\State \textbf{Online Estimation Loop:}
\For{$k = 1, 2, \dots$}
    \Statex \Comment{\textbf{--- Standard \ac{kf} Prediction ---}}
    \State $\hat{\bm{\omega}}_{k|k-1} \gets \hat{\bm{\omega}}_{k-1|k-1}$ {Using random walk model \eqref{eq:randomWalk}}
    \State $\bm{P}_{k|k-1} \gets \bm{P}_{k-1|k-1} + \bm{\Sigma}_{\mathrm{d},k-1}$
    \Statex \Comment{\textbf{--- \ac{kmp}-based Virtual Measurement ---}}
    \State $(\bm{\mu}_{*,k},\bm{\Sigma}_{*,k}) \gets \textsc{KMPPredict}(\bm{\Theta},\bm{s}_*)$ using \eqref{eq:KMP_mean},\eqref{eq:KMP_var}
    \State $\bm{\zeta}^*_{k} \gets \bm{x}_k - \bm{x}_{k-1} - t_\mathrm{s} \bm{u}_{k-1} + t_\mathrm{s}\bm{\mu}_{*,k}$ using \eqref{eq:virtual_measurement}
    \State $   \bm{\Sigma}_{\mathrm{\nu},k} \gets  t_\mathrm{s}^2 \bm{\Sigma}_{*,k} + \bm{\Sigma}_{\mathrm{emp},k}$ using \eqref{eq:noise_cov}
    \State Innovation: $\bm{r}_k \gets \bm{\zeta}^*_k - \bm{H}\hat{\bm{\omega}}_{k|k-1}$.
    \State Update empirical noise: $   \bm{\Sigma}_{\mathrm{emp},k}  \gets  \bm{\Sigma}_{\mathrm{emp},k-1} + \rho \left( \operatorname{diag}(\bm{r}_k^2) - \bm{\Sigma}_{\mathrm{emp},k-1} \right)$ using \eqref{eq:noise_adaptation}
    \Statex \Comment{\textbf{--- \ac{vb} Measurement Update ---}}
    \For{$i = 1$ to $M$}
        \State Compute Kalman gain $\bm{K}_k$ using \eqref{eq:kvark_recursion}.
        \State Update state $\hat{\bm{\omega}}_{k|k}$ and  $\bm{P}_{k|k}$using \eqref{eq:kvark_recursion}.
        \State Update IW parameters $\lambda_{k|k}, \bm{\Upsilon}_{k|k}$ using \eqref{eq:iw_parameter_update}.
        \State Update process noise $\bm{\Sigma}_{\mathrm{d},k}$ using \eqref{eq:sigma_d_VB}.
    \EndFor
    \State $\hat{\bm{\omega}}_{k|k} \gets \hat{\bm{\omega}}_M,\quad \bm{P}_{k|k} \gets \bm{P}_M$
    \State \textbf{Output:} $\hat{\bm{\tau}}_{\mathrm{ext},k} \gets \hat{\bm{\omega}}_{k|k}$
\EndFor
\end{algorithmic}
\end{algorithm}

In the proposed \ac{kvark} framework, this adaptive \ac{vb} is performed at each time step using the innovation of the external torque state. This ensures that the process noise covariance tracks time-varying uncertainty arising from unmodeled dynamics, contact transitions, and \ac{kmp} prediction residuals, while being directly modulated by the measurement uncertainty. The complete algorithm of the proposed framework, \ac{kvark}, is presented in Algorithm \ref{algorithm1}.

\section{Experiments}\label{sec:experiments}
To evaluate the proposed method, we conducted two sets of experiments on a 6-DoF collaborative robotic arm (Orion 5). The first set focused on payload-induced external torques during a dynamic joint-space trajectory; the second involved a physical interaction task and its effect on end-effector Cartesian wrenches. All experiments were implemented and logged in \ac{ros} at 500 Hz. Joint states, commands, and F/T measurements were aligned offline using \ac{ros} timestamps and resampled to a common time grid before evaluation.

\subsection{Experimental Setup}

In both experiments, the ground-truth external effects were measured using a force/torque (F/T) sensor mounted on the end-effector. The F/T sensor used in the experiments was a Shenzhen $\gamma$ 82 series multi-axis force sensor with the signal processor component module Shenzhen D.R 304 with built-in low-pass filtering. The calculations were performed once the data was collected from the experiments on an Ubuntu 20.04 PC with an Intel i7-12700K CPU and 64 GB of RAM. The inertial parameters of the sensor were identified and incorporated into the nominal model $\bm{\tau}_{\mathrm{EL}}$. In the first experiment, the sensor was allowed to measure the payload attached to it. In contrast, in the second experiment, the tool-induced payload effect was compensated through calibration and taring procedures so that the estimated quantities reflected only the task-related interaction forces and torques.

In the first set of experiments, the robot followed dynamically exciting joint-space trajectories under payload effects. More specifically, another dynamically exciting trajectory similar to the one described in \autoref{sec:proposed method} was generated and the robot was tasked to follow this trajectory under position control while a simple weight of 3.2 kg was attached at the end effector. The ground truth external torques were calculated from the mean Cartesian forces and torques recorded by the sensor following the standard mapping:

\begin{equation}
\label{tau_ext_real}
        {\bm{\tau}}_{\mathrm{ext}} = \bm{J}^\top(\bm q)\bm{f}_{\mathrm{ext}}
\end{equation}
where $\bm{J} \in \mathbb{R}^{6\times n}$ is the robot geometric Jacobian defined in the end-effector frame, and $\bm f_{\mathrm{ext}} \in \mathbb{R}^{6\times 1}$ is the Cartesian wrench.

In the second experiment, we equipped the robot with a custom polishing head and flange. The robot executed a square trajectory with flat-surface contact using Cartesian impedance control \cite{mayr2022c++}. \changeOLD{The commanded joint torques under Cartesian impedance control are given by }\footnote{\changeOLD{ We adopted the controller's software structure from \cite{mayr2022c++}, by adding the gravity compensation term  $\bm{g}(\bm{q})$. The code and implementation details can be found in  \url{https://github.com/matthias-mayr/Cartesian-Impedance-Controller}}}

\begin{equation}
\changeOLD{
\begin{aligned}
\bm{\tau}_{\mathrm{c}} &= \bm{\tau}_{\mathrm{c}}^{\mathrm{ca}} + \bm{\tau}_{\mathrm{c}}^{\mathrm{ns}} + \bm{\tau}_{\mathrm{c}}^{\mathrm{ext}} + \bm{g}(\bm{q}), \\
\bm{\tau}_{\mathrm{c}}^{\mathrm{ca}} &= \bm{J}^{\top}(\bm{q})\Big(-\bm{K}_{\mathrm{ca}}\Delta\bm{\xi} - \bm{D}_{\mathrm{ca}}\bm{J}(\bm{q})\dot{\bm{q}}\Big), \\
\bm{\tau}_{\mathrm{c}}^{\mathrm{ns}} &= \Big(\bm{I}_n - \bm{J}^{\top}(\bm{q})\,\bm{J}^{\top\dagger}(\bm{q})\Big)
\Big(-\bm{K}_{\mathrm{ns}}(\bm{q}-\bm{q}_{\mathrm{d}}) - \bm{D}_{\mathrm{ns}}\dot{\bm{q}}\Big), \\
\bm{\tau}_{\mathrm{c}}^{\mathrm{ext}} &= \bm{J}^{\top}(\bm{q})\,\bm{f}_{\mathrm{ext},\mathrm{c}} .
\end{aligned}
}
\end{equation}
\changeOLD{

Here, $\bm{\tau}_{\mathrm{c}}$ is the commanded joint torque, $\bm{q},\dot{\bm{q}}$ are joint position and velocity, and $\Delta\bm{\xi}$ is the Cartesian pose error. $\bm{K}_{\mathrm{ca}},\bm{D}_{\mathrm{ca}}$ and $\bm{K}_{\mathrm{ns}},\bm{D}_{\mathrm{ns}}$ are Cartesian and nullspace stiffness/damping matrices, $\bm{q}_{\mathrm{d}}$ is the desired joint configuration, $\bm{f}_{\mathrm{ext},\mathrm{c}}$ is the commanded Cartesian wrench, and $(\cdot)^{\dagger}$ denotes the Moore--Penrose pseudoinverse. The controller was used with no force modulation i.e. $\bm{\tau}_{\mathrm{c}}^{\mathrm{ext}} = 0$ and with static impedance gains given by,
$
\bm{K}_{\mathrm{ca}} = \operatorname{diag}(K_x, K_y, K_z, K_{R_x}, K_{R_y}, K_{R_z}), 
$
$
\bm{K}_{\mathrm{ns}} = K_{\mathrm{ns}} \bm{I}_n,
$
where $K_x =340\,\mathrm{N/m} \,$, $K_y = 340\,\mathrm{N/m}\,$, $K_z =940\,\mathrm{N/m} ,\,$ $K_{R_x} = 18\,\mathrm{N\cdot m/rad}\,$, $K_{R_y} = 18\,\mathrm{N\cdot m/rad}\,$, $K_{R_z} =18\,\mathrm{N\cdot m/rad} \,$, and $K_{\mathrm{ns}} = 78\,\mathrm{N\cdot m/rad}\,$. The damping matrices $\bm{D}_{\mathrm{ca}}$ and $\bm{D}_{\mathrm{ns}}$ were computed according to the critical damping condition, $\bm{D} = 2\sqrt{\bm{K}}$, applied element-wise to the diagonal stiffness matrices.} 

\begin{figure}[t]
    \centering
      \fontsize{9}{9}
\input{visuals/setup.pdf_tex}%

    \caption{Experimental setup using Orion 5, F/T sensor and custom made polishing tool}
    \label{fig:experiment_setup}
\end{figure}

The measured external wrenches were filtered to obtain ground-truth data for performance evaluation. {Furthermore, since the observers operate in joint space, the inverse of \eqref{tau_ext_real} was employed using a damped Moore–Penrose pseudoinverse to compute the estimated external forces as
\begin{equation}
\hat{\mathbf{f}}_{\mathrm{ext}} 
= \left(\mathbf{J}^\top\right)^{\#}_{\lambda} 
\, \hat{\boldsymbol{\tau}}_{\mathrm{ext}}
\label{f_ext_est}
\end{equation}
 where $(\cdot)^{\#}_{\lambda}$ denotes the damped Moore--Penrose pseudoinverse with damping factor $\lambda$. } The setup of this experiment can be seen in Figure~\ref{fig:experiment_setup}.

\subsection{Residual torque modeling}

To model the residual torques, we collected free-motion  trials of the robot {in its nominal (unloaded) configuration---without the F/T sensor and polishing tool---as described in \autoref{sec:proposed method}. The residual torque was computed per joint and time step using~\eqref{eq:residual_torque}. This dataset isolates the robot’s intrinsic residual dynamics---friction, transmission nonlinearities, and other unmodeled internal effects---independent of external disturbances or payload-induced dynamics. Although a payload can in principle alter frictional or transmission behavior, the unloaded-trained model showed no systematic degradation in residual prediction or external torque estimation in our experiments.

\changeOLD{We collected 10 excitation trajectories of 10 s each, sampled at 500 Hz, resulting in 5,000 samples per joint. Each trajectory consisted of $K = 10$ harmonics with frequencies in [0.1, 2] Hz. The GA used a population of 50 over 100 generations.}
The dataset included joint positions, velocities, accelerations, and torques recorded during a variety of trajectories. From this dataset, $20\%$ of the samples were randomly selected as a test set, ensuring that evaluation was performed on previously unseen data. \changeOLD{Furthermore, this dataset was constructed so that both training and test partitions were not exposed to any of the evaluation trajectories used in Experiments 1 and 2.}
Figure~\ref{fig:gmr_ref} illustrates the \ac{gmr} reference residual torque trajectories used to initialize both the \ac{gmr}-\ac{gp} and \ac{kmp} models. The black lines represent the mean residual torque as a function of joint velocity, and the shaded areas depict variance. The blue dots correspond to the held-out test data.

\begin{figure}[t]
  \centering
  \def\svgwidth{1\linewidth}
  \fontsize{9}{9}
  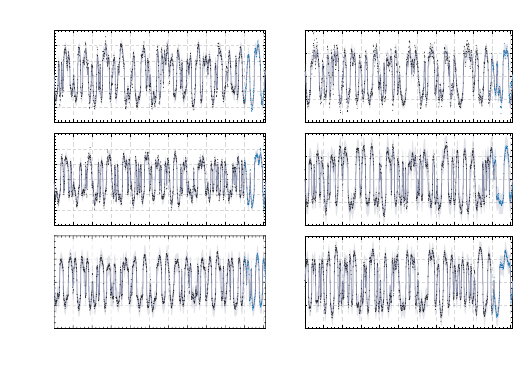%
  
\caption{\ac{gmr} reference trajectories for residual torque modeling across six joints. The black curves indicate the \ac{gmr}-predicted mean and variance; blue dots represent test data.}
\label{fig:gmr_ref}
\end{figure}

To capture the structure in the residual torque data, we fitted Gaussian components using the \ac{em} algorithm on the velocity-residual torque pairs. Figure~\ref{fig:gmm_vel_residual} shows the scatter plots of velocity versus residual torque for each joint, overlaid with the Gaussian components learned by \ac{em}. These models form the probabilistic foundation for the subsequent \ac{gmr}, \ac{gmr}-\ac{gp}, and \ac{kmp} modeling.

\begin{figure}[t!]
    \centering
    \def\svgwidth{1\linewidth}
    \fontsize{9}{9}
    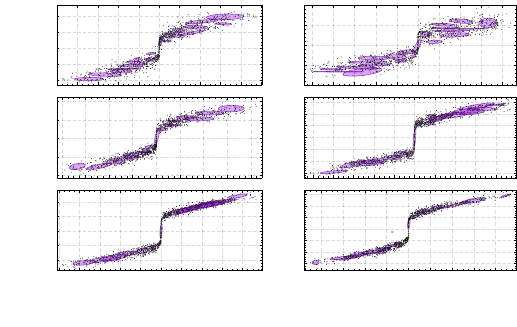%
  
    \caption{Velocity versus residual torque scatter plots for each joint with Gaussian components fitted by the \ac{em} algorithm. The components capture the nonlinear frictional behavior across the velocity range.}
    \label{fig:gmm_vel_residual}
\end{figure}

We evaluated three state-of-the-art observer formulations to compare our proposed algorithm:

\begin{itemize}[leftmargin=*]
    \item \textbf{\ac{gmr}-\ac{gp}:} Our combined Gaussian Mixture Regression and Gaussian Process method. \ac{gmr} produces the reference {residual-torque} distribution, and the \ac{gp} {is trained} on sampled data from this distribution, effectively downsampling the dataset {while preserving the relevant information}. {The \ac{gmr} reference trajectory used 20 components, identical to the one used for \ac{kvark}. The \ac{gp} used a squared-exponential kernel with \ac{ard}-based length-scale selection. The noise/regularization parameter was selected consistently with the \ac{kmp} parameter $\lambda_1$, so that the mean residual-torque predictions of \ac{gp} and \ac{kmp} were compared under matched regularization settings.}

    \item \textbf{\ac{kvark}:} Our proposed Kernelized Movement Primitive approach is initialized with the same {20-component \ac{gmr}} reference trajectory. {The squared-exponential kernel length-scale $l$ was obtained using \ac{ard}-based kernel hyperparameter optimization. The mean-regularization/noise-level parameter $\lambda_1$ was selected from the optimized likelihood/noise level and shared with the \ac{gp}-based model to keep the mean residual-torque prediction comparable. The \ac{kmp}-specific covariance-scaling parameter $\lambda_2$ was tuned empirically to obtain meaningful heteroscedastic uncertainty. The final hyperparameters used were:}
    \[
    \begin{aligned}
    l &= [0.0100, \; 0.0200, \; 0.0110, \; 0.0790, \; 0.1154, \; 0.1244], \\
    \lambda_{1} &= [0.0676, \; 0.0020, \; 0.5032, \; 0.0729, \; 0.0300, \; 0.03418], \\
    \lambda_{2} &= 10^{5} \times [10, \; 20, \; 20, \; 20, \; 20, \; 20], \\
    \sigma_f^2 &= 10^{4}.
    \end{aligned}
    \]

    \item \textbf{\ac{gpadkf}:} {The sparse Gaussian Process Adaptive Disturbance Kalman Filter baseline follows the formulation described in \cite{wei2023contact}. It models the residual torque using a sparse \ac{gp} and incorporates the \ac{gp} predictive covariance into the adaptive Kalman filter. We used a squared-exponential \ac{ard} kernel with 50 inducing points, and the remaining settings followed the reference formulation with adaptations only for the Orion 5 data format and sampling rate.}

    \item \textbf{\ac{nnadkf}:} Similar to the approaches in \cite{liu2018end} and \cite{scholl2024learning}, a neural network was trained to estimate the mean residual torque, which was then used in the Kalman-filter virtual measurement. {One network was trained per joint. The base network consisted of two hidden layers with 30 ELU neurons each and was trained using Adam with learning rate $10^{-3}$, 300 epochs, and mini-batch size 512. The additive correction network consisted of one hidden layer with 30 ELU neurons and was trained using Adam with learning rate $10^{-2}$, 5 epochs, and mini-batch size 128. Since the neural network estimates only the mean residual torque and does not provide predictive uncertainty, a static covariance matrix was used.}
\end{itemize}

{For all Kalman-filter-based observers, including the static covariance used in \ac{nnadkf} and the initial covariances on adaptive variants, the required values were selected using the covariance-matching procedure described in Section~\ref{sec:ablation}. While the numerical values reported above are platform-specific rather than universal, the same tuning procedure can be applied to other manipulators.}

Figure \ref{fig:residual_estimation} illustrates the residual torque predictions on the test data. Each subplot shows the true residual torques (black dots), along with the mean predictions and $\pm 2\sigma$ confidence intervals from the four models. 
\begin{figure}[t]
  \centering
  \def\svgwidth{1\linewidth}
  \fontsize{9}{9}
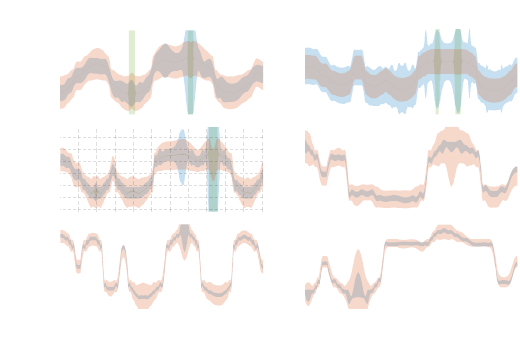%

  \caption{Residual torque estimates on the test joints for \ac{gpr} (light green), \ac{kmp} (orange), Sparse \ac{gp} (blue) Neural Networks (Dark Green) together with their confidence intervals}
   \label{fig:residual_estimation}  

\end{figure}
  
Quantitative evaluation was performed using the \ac{rmse} computed over the test set. The joint-wise \ac{rmse} values are given in Table~\ref{tab:joint-wise-rmse}.
\begin{table*}[t]
 \caption{\ac{rmse} of residual torque estimation comparison across methods for each joint.}
\centering
\begin{tabular}{|c|c|c|c|c|}
\hline
\textbf{Joint} & \textbf{\ac{rmse} (Sparse \ac{gp}) [N$\cdot$m]} & \textbf{\ac{rmse} (\ac{gmr}-\ac{gp}) [N$\cdot$m]} & \textbf{\ac{rmse} (\ac{kmp}) [N$\cdot$m]} & \textbf{\ac{rmse} (\ac{nn}) [N$\cdot$m]} \\
\hline
1 & 1.4445 & 1.5333 & 1.5071 & 2.1415 \\
2 & 3.5483 & 2.7168 & 2.7322 & 2.812\\
3 & 1.9966 & 1.4498 & 1.4430 & 1.0622\\
4 & 0.3256 & 0.3311 & 0.3245 & 0.3045\\
5 & 0.2424 & 0.2866 & 0.2737 & 0.2476\\
6 & 0.2350 & 0.2685 & 0.2586 & 0.2405\\
\hline
\end{tabular}
\label{tab:joint-wise-rmse}

\end{table*}

These results are summarized in Figure~\ref{fig:rmse_bar}. Both \ac{gmr}-\ac{gp} and \ac{kmp} achieved similar accuracy, with \ac{kmp} showing smoother variance behavior and better uncertainty calibration in regions with sparse data. The sparse \ac{gp} baseline exhibited higher error, particularly in joints with more complex dynamical and frictional behavior. Furthermore, \ac{kmp} successfully captured the naturally occurring variance in the residual torque data, whereas the other methods captured only the uncertainty, and neural networks, in particular, conveyed little to no information about the underlying variance.
\begin{figure}[t!]
  \def\svgwidth{\linewidth}
  \fontsize{9}{9}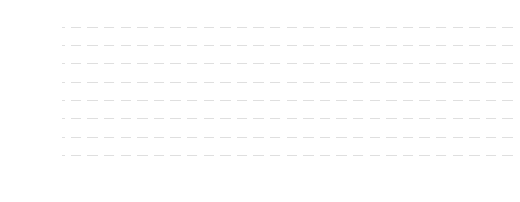
      \caption{Average RMSE of residual torque \changeOLD{modeling} for 4 different methods.}
      \label{fig:rmse_bar}
\end{figure}

Overall, \ac{kmp} demonstrated an advantage in providing reliable mean and variance predictions, which is critical for integration into uncertainty-aware observers such as the adaptive disturbance Kalman filter used in this study.

\subsection{External force estimation}
We evaluated the performance of \ac{gmr}-\ac{gp}, \ac{kmp} and baseline observers in estimating the external joint torques {and Cartesian forces for both experiments.}
\subsubsection{Experiment 1}
 {In this experiment,} the robot executed joint-space trajectories {with a payload attached to the end effector that is unknown to the estimators}. The true external torques were {calculated} as in \eqref{tau_ext_real}.
The task of the observers was to estimate these external torques without direct force-torque sensing, using only the robot's internal measurements and the residual torque models integrated into the adaptive Kalman filter.

Figure~\ref{fig:ext_torque_timeseries} shows the time histories of { the external torque estimation errors} by each method. Both the \ac{nnadkf} and \ac{kmp} observers closely track the true external torques, while the sparse \ac{gp} baseline exhibits larger deviations and higher-frequency fluctuations.

\begin{figure}[t]
    \centering
    \def\svgwidth{1\linewidth}
    \fontsize{9}{9}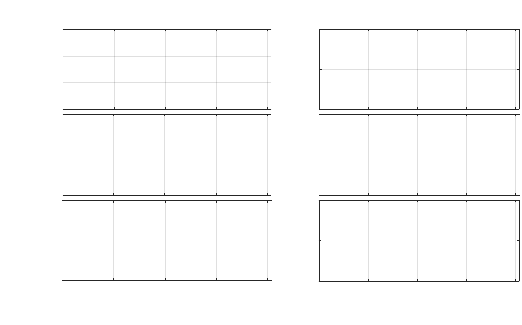%
  
    \caption{{Time histories of external torque estimation errors for all six joints on experiment 1: comparison of errors from \ac{gmr}-\ac{gp} (blue), \ac{kvark} (red), \ac{gpadkf} (green) and \ac{nnadkf} (light blue)}}
    \label{fig:ext_torque_timeseries}
\end{figure}

Quantitative results, summarized in Figure~\ref{fig:ext_torque_rmse}, show that the \ac{kmp}-based observer achieved the lowest average RMSE across joints, followed by {\ac{nnadkf}}. The sparse \ac{gp} baseline produced the highest error.

\begin{figure}[t]
    \centering
  \def\svgwidth{1\linewidth} 
  \fontsize{9}{9}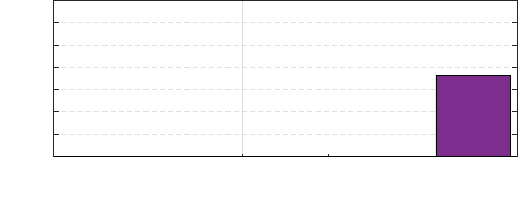%
    \caption{{Average RMSE of external torque estimations with standard deviations for the four algorithms.}}
    \label{fig:ext_torque_rmse}
\end{figure}

\subsubsection{{Experiment 2}}
The second experiment also yielded similar results. The joint-space and Cartesian-space external forces and their estimates can be seen in Figure~\ref{fig:exp_sensor_cart}.

\begin{figure}[t!]
    \centering
  \def\svgwidth{1\linewidth}
  \fontsize{9}{9}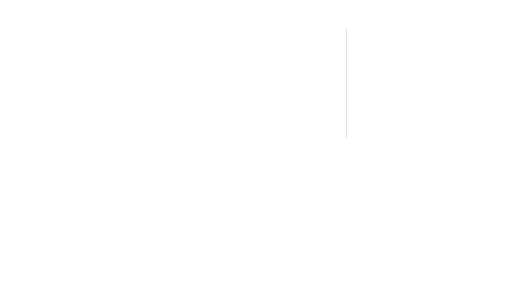%
    \caption{\changeOLD{Errors on external Force Estimates for 6 Cartesian Directions on experiment 2}}
    \label{fig:exp_sensor_cart}
\end{figure}

Table~\ref{tab:rmse_combined} reports the joint-space and Cartesian-space RMSE values for all four methods. \ac{kvark} achieves the lowest average Cartesian force error (6.76~N), followed by \ac{gmr}-\ac{gp} (7.14~N), \ac{nnadkf} (7.45~N), and \ac{gpadkf} (7.83~N). \ac{kvark} also yields the lowest joint-space torque RMSE, while moment estimation errors are comparable across \ac{gmr}-\ac{gp}, \ac{kvark}, and \ac{gpadkf}.

\setlength{\tabcolsep}{4.5pt} 

\begin{table*}[t]
\centering
\caption{{Comparison of RMSE values for joint-space torques $\tau_i$ [N$\cdot$m] and Cartesian-space forces and moments $(F_x, F_y, F_z, m_x, m_y, m_z)$ in experiment 2. Average RMSEs for both domains are shown as mean $\pm$ standard deviation across joints or Cartesian axes. The lowest value in each column is highlighted in bold.}}
\label{tab:rmse_combined}
\resizebox{\textwidth}{!}{
\begin{tabular}{|c|cccccc|cccccc|c|c|c|}
\hline
\textbf{Method}
& $\tau_1$ & $\tau_2$ & $\tau_3$ & $\tau_4$ & $\tau_5$ & $\tau_6$
& $F_x$ & $F_y$ & $F_z$ & $m_x$ & $m_y$ & $m_z$
& \textbf{Joint} & \textbf{Force} & \textbf{Moment} \\
\hline
GMR-GP
& 2.73 & 4.83 & 3.01 & 1.94 & 0.86 & 0.82
& 5.83 & 10.17 & 5.42 & 1.70 & 0.62 & 0.62
& 2.36 $\pm$ 1.52 & 7.14 $\pm$ 6.21 & 0.98 $\pm$ 0.94 \\

K-VARK
& \textbf{2.73} & \textbf{4.56} & \textbf{2.63} & 1.98 & 0.92 & 0.76
& 5.77 & \textbf{9.89} & \textbf{4.61} & \textbf{1.67} & 0.67 & 0.56
& \textbf{2.26 $\pm$ 1.40} & \textbf{6.76 $\pm$ 5.72} & 0.97 $\pm$ 0.93 \\

GPADKF
& 2.78 & 5.02 & 5.01 & \textbf{1.84} & \textbf{0.84} & \textbf{0.59}
& \textbf{5.51} & 10.10 & 7.87 & 1.78 & \textbf{0.66} & \textbf{0.43}
& 2.68 $\pm$ 1.97 & 7.83 $\pm$ 6.47 & \textbf{0.95 $\pm$ 0.88} \\

\ac{nnadkf}
& 3.14 & 5.00 & 3.80 & 2.42 & 0.86 & 0.94
& 6.52 & 10.54 & 5.28 & 2.13 & 0.67 & 0.69
& 2.69 $\pm$ 1.63 & 7.45 $\pm$ 6.87 & 1.16 $\pm$ 1.07 \\
\hline
\end{tabular}
}
\end{table*}

{The remaining mismatch in Cartesian wrench estimation is mainly due to contact effects that are only imperfectly captured by the rigid-body model and Jacobian mapping, including tool compliance, impedance-control closed-loop dynamics, and surface friction. These effects are most pronounced in the tangential directions, where stiction/micro-slip and friction-induced disturbances dominate the interaction, leading to larger errors than in the normal direction. This explains why the largest discrepancies are observed primarily along the tangential axes during the surfacing task.}
\changeOLD{Table~\ref{tab:cartesian_robustness_final} reports additional robustness-oriented statistical metrics in Cartesian space. In addition to mean absolute error, we evaluate tail behavior (95th percentile), worst-case deviation, and temporal stability of the estimation error. K-VARK yields the lowest values across these metrics, indicating reduced high-percentile deviations and improved stability under contact conditions. These results suggest that the performance differences observed in the average-error analysis are consistent with improvements in robustness and temporal behavior.}

\begin{table}[t]
\centering
\caption{\changeOLD{Cartesian robustness metrics for experiment 2. 
Mean Absolute Error captures central tendency, P95 reflects tail robustness, 
Max denotes worst-case deviation, and Temporal STD measures stability over time. 
Lower values indicate better performance.}}
\begin{tabular}{|c|cccc|}
\hline
\textbf{Method}
& \textbf{Mean Abs.}
& \textbf{P95}
& \textbf{Max}
& \textbf{Temp. STD} \\
\hline
GMR-GP 
& 3.25 & 12.09 & 27.75 & 3.58 \\
K-VARK 
& \textbf{3.14} & \textbf{11.53} & \textbf{23.46} & \textbf{3.32} \\
GPADKF 
& 3.57 & 13.24 & 37.25 & 3.68 \\
\ac{nnadkf} 
& 3.32 & 13.19 & 34.52 & 3.97 \\
\hline
\end{tabular}
\label{tab:cartesian_robustness_final}
\end{table}

In addition to accuracy, computational efficiency was assessed. Figure~\ref{fig:comp_time} presents the total and per-sample processing times. The \ac{kmp}-based observer demonstrated the best trade-off between accuracy and speed, with lower computational cost than both \ac{gmr}-\ac{gp} and the sparse \ac{gp} baseline.

\begin{figure}[t]
    \centering
    \def\svgwidth{1\linewidth}
    \fontsize{9}{9}
    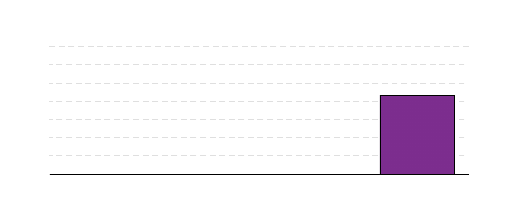
    \caption{Comparison of total computation time (left axis) and average processing time per sample (right axis) for GMR-GP, GPADKF, K-VARK, and \ac{nnadkf}.}
    \label{fig:comp_time}
\end{figure}

These results demonstrate that \ac{kmp} not only provides accurate external torque estimates but also maintains computational efficiency suitable for real-time implementation in sensorless force estimation scenarios.

\subsection{\changeOLD{Ablation Studies}} \label{sec:ablation}

\changeOLD{ In order to assess the contribution of each component in \ac{kvark}, a structured ablation suite was prepared and conducted on both experiments. The same residual torque models and data splits were used across all methods. We consider five observer configurations with progressively increasing modeling and adaptation complexity which are given in~\autoref{tab:ablation_components_compact}.}

\begin{table}[t]

\caption{\changeOLD{Ablation variants and which component are enabled.}}
\label{tab:ablation_components_compact}
\centering
\setlength{\tabcolsep}{3.5pt}
\renewcommand{\arraystretch}{1.05}
\begin{tabular}{@{}lccc@{}}
\toprule
Variant & $\Sigma^*(\dot{q})$ & EWMA $\Sigma_{\mathrm{emp},k}$ & $\Sigma_{d,k}$ update \\
\midrule
A1: Standard KF                        & --         & --         & -- \\
A2: KMP-KF                     & \checkmark & --         & -- \\
A3: + EWMA-$R_k$                             & \checkmark & \checkmark & -- \\
A4: + Innov.\ scaling ($\alpha_k \Sigma_{d,0}$) & \checkmark & \checkmark & \checkmark \\
A5: K-VARK (\ac{vb}-IW on $\Sigma_{d,k}$)         & \checkmark & \checkmark & \checkmark \\
\bottomrule
\end{tabular}
\end{table}

\paragraph{A1: Standard KF}
\changeOLD{A classical disturbance Kalman filter with fixed process and measurement covariances $(\Sigma_d,\Sigma_\nu)$ tuned offline}.

\paragraph{A2: KMP-KF}
\changeOLD{Incorporates the KMP-predicted heteroscedastic measurement variance $\Sigma^*(\dot{q})$ but keeps $\Sigma_d$ and $\Sigma_{\text{emp}}$ fixed.}

\paragraph{A3: KMP-KF + EWMA}
\changeOLD{Adds exponential adaptation of $\Sigma_{\text{emp},k}$ using the innovation statistics as in~\eqref{eq:noise_adaptation}, while $\Sigma_d$ remains fixed.}

\paragraph{A4: KMP-KF + innovation scaling}
\changeOLD{Uses the same measurement model as above but replaces \ac{vb} adaptation with a simple innovation-based scaling of the process covariance:}
\[
\changeOLD{
\Sigma_{d,k} = \alpha_k \Sigma_{d,0},
}
\]
\changeOLD{
where $\alpha_k$ is updated from the \ac{nis}, providing a lightweight adaptive baseline.}

\paragraph{A5: K-VARK}
\changeOLD{Activates all components: heteroscedastic $\Sigma^*(\dot{q})$, empirical measurement adaptation, and \ac{vb}-based recursive estimation of $\Sigma_{d,k}$.}

\changeOLD{For a fair comparison across ablations, all covariance terms that remain static in a given variant were identified from the training dataset, and the same values were used as initial covariances for the adaptive baselines. For ablations without heteroscedastic variance, a constant measurement covariance was defined as}

\changeOLD{\[
\Sigma^*_{\text{const}}
=
\mathrm{diag}\!\left(
\mathrm{median}\{\sigma^*_j(\dot{q})\}
\right),
\]}
\changeOLD{where $\sigma^*_j(\dot{q})$ denotes the predicted measurement variance  evaluated using classical innovation-based covariance matching over the training data. The fixed process covariance $\Sigma_{d,\text{const}}$ was tuned via the same strategy on the training trajectory by scaling an initial $\Sigma_{d,0}$ such that the empirical mean \ac{nis} matched its theoretical expectation. The empirical measurement covariance $\Sigma_{\text{emp},\text{const}}$ was estimated from the median squared innovation during training, yielding a diagonal covariance that captures stationary sensor and modeling noise. These values were kept fixed in non-adaptive ablations and used as initialization for the adaptive variants.}

\subsubsection{Experiment 1}
\changeOLD{In Experiment~1, we analyze the complete trajectory under a payload that is unknown to the estimator. ~\autoref{fig:abl_1_var} shows the time histories of the covariances for all ablation variants, while \autoref{tab:abl_avg_rmse} shows the \ac{rmse} values of the estimations.}

\begin{figure}[t!]
    \centering
  \def\svgwidth{1\linewidth}
  \fontsize{9}{9}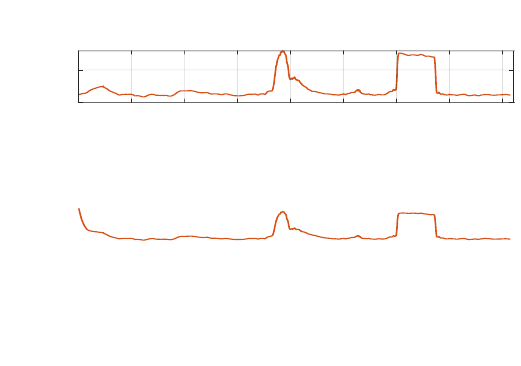%
    \caption{\changeOLD{Time histories of traces of covariances for different variations on experiment~1, with traces of covariances on y-axis}}
    \label{fig:abl_1_var}
\end{figure}

\begin{table}[t]
\caption{\changeOLD{Experiment 1 – Average RMSE across joints over full trajectory.}}
\label{tab:abl_avg_rmse}
\centering
\setlength{\tabcolsep}{8pt}
\renewcommand{\arraystretch}{1.05}
\begin{tabular}{@{}lc@{}}
\toprule
Variant & Avg RMSE [N$\cdot$m] \\
\midrule
Standard KF                        & 2.832 \\
KMP-KF                      &  2.532 \\
KMP-KF + $\Sigma_{\mathrm{emp},k}$ update & 2.014 \\
KMP-KF + Innov.\ scaling                  & 1.669 \\
K-VARK                     & 1.439 \\
\bottomrule
\end{tabular}
\end{table}

The full-trajectory analysis in Experiment~1 shows how uncertainties evolve during operation. The input-dependent measurement covariance varies systematically along the trajectory: in well-sampled velocity regions it preserves the heteroscedastic variance structure of the training data, reflecting the intrinsic variability of the residual dynamics, while in sparsely sampled regions it increases to signal reduced confidence under limited data support. The uncertainty thus does not act as a uniform inflation term, but retains the learned variance profile where data is dense and expands only where extrapolation occurs. The empirical measurement covariance adapts gradually through the innovation statistics, compensating for slow variations in sensor and modeling noise. For the process model, both the innovation-scaling baseline and the \ac{vb} update adapt $\Sigma_{d,k}$, but the \ac{vb} variant does so in a structured way that also responds to the measurement-covariance changes induced by the other adaptive components. Quantitatively, the RMSE over the full motion decreases consistently as components are activated, reaching its lowest value for the K-VARK configuration, where structured measurement uncertainty and adaptive process covariance act jointly.

\subsubsection{Experiment 2}
In Experiment~2, we focus the ablation study during the initial contact transient phase, where the influence of each component is most significant. Figure~\ref{fig:abl2_var} shows the time evolution of the force estimation error along the contact axis for all variants, and Table~\ref{tab:abl_exp2_rmse} summarizes the average error in the same window. Since the robot motion changes only mildly during this short interval, the operating point remains within a narrow velocity range; consequently, the input-dependent measurement covariance stays nearly constant throughout the transient. This explains why the standard fixed-covariance KF and the heteroscedastic-measurement variant yield comparable errors in this segment showing that the transient is dominated by disturbance dynamics rather than measurement reweighting.

\begin{table}[t]
	\caption{{Experiment 2 – Average RMSE and Peak Error During Contact Transient on Contact Axis (z)}}
	\label{tab:abl_exp2_rmse}
	\centering
	\setlength{\tabcolsep}{8pt}
	\renewcommand{\arraystretch}{1.05}
	\begin{tabular}{@{} l c c @{}}
		\toprule
		Variant & Avg RMSE [N] & Peak Error [N] \\
		\midrule
		Standard KF                        & 29.22 & 42.72 \\
		KMP-KF                      &  29.52 & 43.05\\
		KMP-KF + $\Sigma_{\mathrm{emp},k}$ update & 42.26 & 58.29\\
		KMP-KF + Innov.\ scaling                  &19.42  &35.42 \\
		K-VARK                     & 7.16 &11.50 \\
		\bottomrule
	\end{tabular}
\end{table}

Interestingly, using empirical measurement-noise adaptation alone degrades performance in this regime. This behavior is consistent with the innovation-driven update: during contact onset, the innovation energy rises sharply due to a sudden unmodeled external wrench. If only $\Sigma_{\mathrm{emp},k}$ is adapted, the filter primarily interprets this innovation burst as increased measurement noise, inflating $\Sigma_{\nu,k}$ and effectively down-weighting the virtual measurement right when rapid correction is needed. As a result, the filter becomes overly conservative and lags the contact transient, leading to larger peak and integrated errors. When a process-noise adaptation mechanism is enabled, the same innovation burst is instead attributed (at least in part) to a change in the disturbance dynamics, increasing $\Sigma_{d,k}$ and allowing faster state variation; accordingly, combining empirical adaptation with process-noise updates improves the transient response.

Finally, the \ac{vb}-based variant achieves the lowest error during contact onset. The difference between innovation scaling and \ac{vb} can be attributed to the fact that both the empirical measurement adaptation and the innovation-scaling baseline are driven directly by the measurement innovation, reacting to the same discrepancy between predicted and observed measurements. In contrast, the \ac{vb} scheme adapts the process covariance based on the estimated change in the external torque state itself, i.e., how much the state must evolve to explain the data. This distinction separates direct innovation-driven adjustment from state-driven covariance adaptation. Moreover, the \ac{vb} update modifies the full covariance structure in a regularized manner, rather than applying a scalar scaling factor. During contact transients, where the external torque undergoes a rapid change, this structured and state-consistent adaptation of process uncertainty yields improved transient tracking performance.

The computational demand of the ablation variants was also examined under the same implementation and hardware. As expected, the standard fixed-covariance \ac{kf} was the fastest variant; however, adding the \ac{kmp} query and adaptive covariance updates resulted in a modest overhead, with the full \ac{kvark} implementation increasing the per-sample processing time by approximately 20\% relative to the standard \ac{kf}.

\begin{figure}[t!]
    \centering
  \def\svgwidth{1\linewidth}
  \fontsize{9}{9}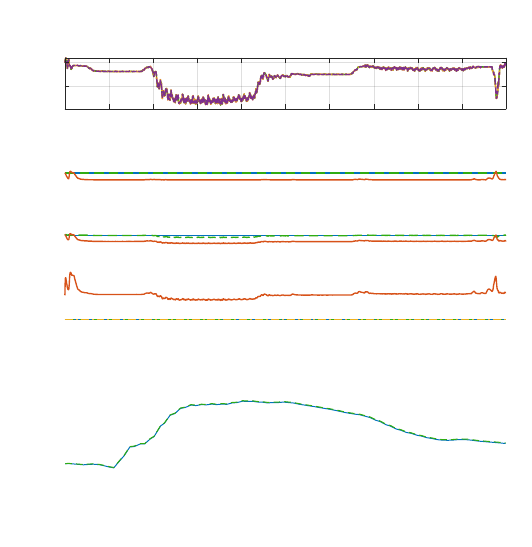%
    \caption{{Time histories of traces of covariances for different variations on experiment~2, with traces of covariances on y-axis and focus on the initial contact transient errors on the axis of contact. Vertical lines represent the initial contact and stabilization respectively}}
    \label{fig:abl2_var}
\end{figure}

\section{Discussion} \label{sec:Discussion}
While developing the \ac{kmp} pipeline for residual torque estimation,  a high-dimensional input space composed of joint positions ($\bm{q}$), velocities ($\dot{\bm{q}}$), and accelerations ($\ddot{\bm{q}}$), {was} employed, resulting in an $n \times 3$ feature matrix for $n$ joints. However, empirical investigations revealed significant generalization issues when using \ac{gmm}s trained with  \ac{em} in such high-dimensional settings.

Specifically, the \ac{em} algorithm tends to overfit sparse regions in the feature space, causing the Gaussian components to shrink excessively~\cite{adams2019survey,yao2025bayesian}. This effect results in poor extrapolation, as the Gaussians fail to represent unobserved yet plausible regions of the input space. Figure~\ref{fig:GMR_overfitted} illustrates this phenomenon using 40 Gaussian components across 18 features, where regression performance is constrained to training trajectories, with substantial degradation on unseen test data.

\begin{figure}[t]
    \centering
    \def\svgwidth{1\linewidth}
    \fontsize{9}{9}
    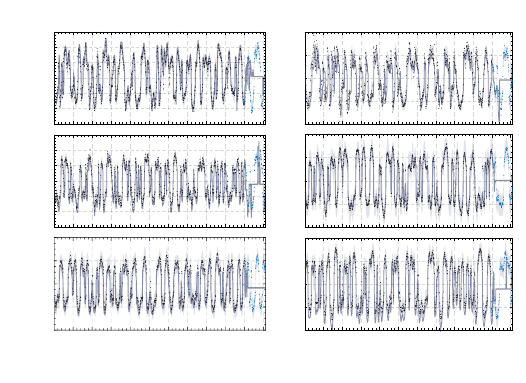%
\caption{\ac{gmr} fitting with \ac{em}. Blue dots represent unseen (test) data points}
\label{fig:GMR_overfitted}
\end{figure}

To address this, we hypothesized that a lower-dimensional representation might yield better generalization without a major compromise in prediction fidelity. Guided by both empirical evidence and physical reasoning, we explored a feature-restricted model using only joint velocity $\dot{q}_i$ as the input for predicting the residual torque $\tau_{\mathrm{r},i}$ of joint $i$. This restriction is motivated by two key observations:

\begin{itemize}
    \item First, in an ideal setting with accurate dynamic modeling, the residual torques should primarily arise from frictional forces, which are well-characterized by models dependent on joint velocity (e.g., Coulomb and Stribeck friction).
    \item Second, \ac{gmm} inherently models heteroscedasticity in the data. In the one-dimensional input space, the variance captured by the mixture components implicitly reflects dependencies on omitted features such as joint position or acceleration. That is, even though the regression mean is modeled as a function of velocity alone, the conditional variance encodes uncertainty due to latent dependencies.
    
\end{itemize}

Figure~\ref{fig:residual_vs_velocity} plots residual torque versus joint velocity across all joints, revealing a strong friction-like dependency. In particular, joints 1--3 exhibit higher variance, likely from larger dynamic effects or sensor noise. 
\begin{figure}[t]
    \centering
    \def\svgwidth{1\linewidth}
    \fontsize{9}{9}
    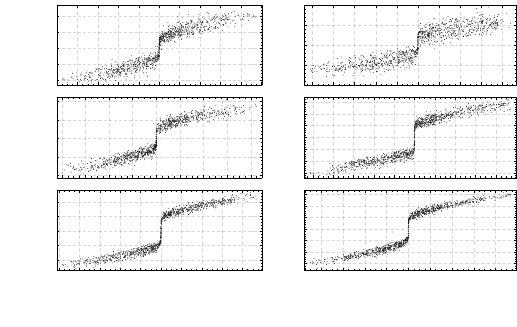%
\caption{Residual torque plots against velocity for all joints}
\label{fig:residual_vs_velocity}
\end{figure}

We assess feature relevance using an \ac{ard} kernel. The log-scale plot (Figure~\ref{fig:ARD_kernel_relevance}) shows velocity as the dominant predictor across joints; other features contribute only marginally.

\begin{figure}[t]
    \centering
    \def\svgwidth{1\linewidth}
    \fontsize{9}{9}
    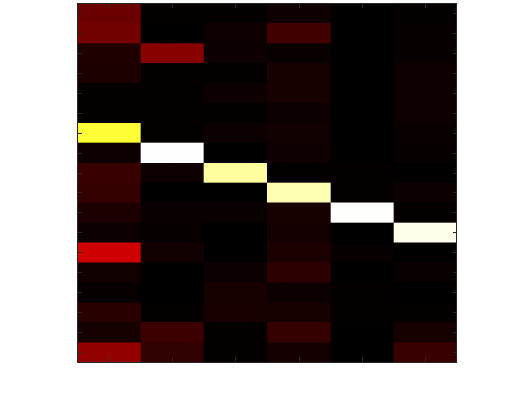%
\caption{\ac{ard} Kernel heatmap for the features}
\label{fig:ARD_kernel_relevance}
\end{figure}

From a practical perspective, employing a lower-dimensional feature space mitigates the curse of dimensionality and allows more effective utilization of available data. Reducing the number of input features enables the use of more Gaussian components without overfitting, enhancing the model's flexibility and robustness.
\subsection{{Internal vs. External Ground-Truth Construction}}

While Section~\ref{sec:experiments} evaluates the observers against an external ground truth measured by an F/T sensor, we additionally report an \emph{internal} ground-truth construction that does not require any external sensing. This is intended to help practitioners validate the proposed framework even when an F/T sensor is unavailable. More specifically, for each experiment, we executed the \emph{same reference trajectory twice}: once with the external effect present (payload attached in Experiment~1, contact with the workpiece in Experiment~2), and once in a corresponding \emph{reference-free} condition (no payload in Experiment~1, no contact/workpiece in Experiment~2). Denoting the commanded motor torques in these two runs by $\bm{\tau}_{\mathrm{m}}^{\mathrm{effect}}$ and $\bm{\tau}_{\mathrm{m}}^{\mathrm{ref}}$, we define the internally constructed external torque as
\begin{equation}
\label{eq:tau_ext_internal_gt}
\bm{\tau}_{\mathrm{ext,int}}
\;\coloneqq\;
\bm{\tau}_{\mathrm{m}}^{\mathrm{effect}}
-
\bm{\tau}_{\mathrm{m}}^{\mathrm{ref}}.
\end{equation}

Intuitively, \eqref{eq:tau_ext_internal_gt} treats the additional torque required to track the same motion as the contribution of the external effect. This methodology to infer the payload effects from internal measurements is widely used in payload identification methods such as \cite{kurdas2022online}.\\
It should be noted, however, that this internal construction is only valid to the extent that the two runs are comparable. In particular, it relies on (i) identical controller gains and structure, (ii) comparable tracking errors and phase delays, (iii) consistent torque bias/offsets, and (iv) stable drive gains (torque constants). In practice, discrepancies such as inaccurate drive gains introduce systematic scaling errors in $\bm{\tau}_{\mathrm{m}}$ and therefore directly corrupt $\bm{\tau}_{\mathrm{ext,int}}$. Therefore, we advise the practitioners who plan to validate this work to use drive gain calibration methodologies such as~\cite{gautier2014global}.\\
Figures~\ref{fig:ext_torque_timeseries_int} and~\ref{fig:exp_sensor_cart_int} report the same evaluation plots as in Section~\ref{sec:experiments}, but using the internal ground truth. Table~\ref{tab:rmse_algorithms} summarizes average RMSE for both ground-truth definitions.

Across both experiments, it can be seen that the ranking of methods is consistent, in particular, \ac{kvark} achieves the lowest error under both ground-truth constructions. However, we observe that the agreement between $\bm{\tau}_{\mathrm{ext,ext}}$ (sensor-based ground truth) and $\bm{\tau}_{\mathrm{ext,int}}$ (internal ground truth) is \emph{tighter} in Experiment~1 than in Experiment~2. This difference can be explained to some degree by taking into account that Experiment~1 is dominated by smooth trajectory and payload effects under position control, where the two-run comparability assumptions are more likely to hold. In contrast, Experiment~2 uses Cartesian impedance control in contact, where several additional factors break the symmetry between the ``effect'' and ``reference'' runs. These include contact-dependent tracking deviations, frictional and stiction effects at the tool-surface interface, time synchronization issues, compliance and deformation in the tool/workpiece, and closed-loop impedance dynamics that can change the commanded torques even at identical reference motion. As a result, the torque difference in \eqref{eq:tau_ext_internal_gt} captures not only the external wrench contribution, but also contact-induced changes in control effort and unmodeled interaction dynamics. \\
Overall, the internal ground truth provides a useful, sensor-free \emph{validation proxy}, particularly for non-contact or payload-type evaluations (as in Experiment~1). For contact-rich impedance-controlled tasks (as in Experiment~2), it should be interpreted more cautiously as an \emph{interaction-induced torque difference} rather than a pure external wrench ground truth. Importantly, despite these limitations, both ground-truth constructions lead to the same qualitative conclusion regarding the relative performance of the compared observers.

\begin{figure}[t]
    \centering
    \def\svgwidth{1\linewidth}
    \fontsize{9}{9}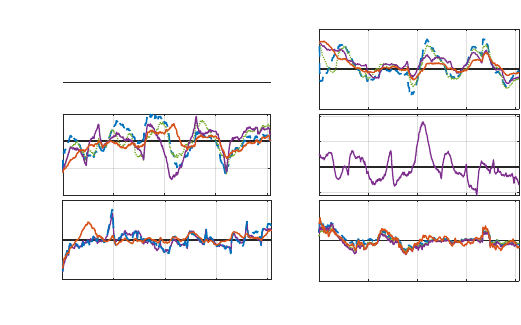%
  
    \caption{{Time histories of external torque estimation errors for all six joints on experiment 1 with internal measurements: comparison of errors from \ac{gmr}-\ac{gp} (blue), \ac{kvark} (red), \ac{gpadkf} (green) and \ac{nnadkf} (light blue)}}
    \label{fig:ext_torque_timeseries_int}
\end{figure}

\begin{figure}[t!]
    \centering
  \def\svgwidth{1\linewidth}
  \fontsize{9}{9}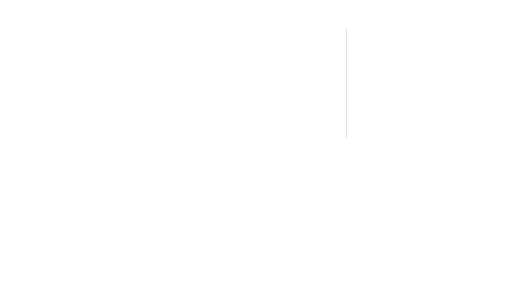%
    \caption{{Errors on external Force Estimates for 6 Cartesian Directions on experiment 2 with internal measurements}}
    \label{fig:exp_sensor_cart_int}
\end{figure}

\begin{table}[]
\centering
\caption{{Average RMSE comparison under two ground-truth constructions: external sensor-based ground truth (Ext-GT) and internal torque-difference ground truth (Int-GT) defined in~\eqref{eq:tau_ext_internal_gt}}}
\label{tab:rmse_algorithms}
\setlength{\tabcolsep}{4pt}
\renewcommand{\arraystretch}{1.05}
\begin{tabular}{lcccc}
\toprule
\textbf{Metric} & \textbf{GMR--GP} & \textbf{GPADKF} & \textbf{K-VARK} & \textbf{\ac{nnadkf}}\\
\midrule
\multicolumn{5}{l}{\textit{Joint torques}}\\
Avg RMSE (Ext-GT) & 2.1968 & 2.4362 & \textbf{1.4393} & 1.9082\\
Avg RMSE (Int-GT) & 2.1804 & 2.3020 & \textbf{1.3962} & 1.8690\\
\midrule
\multicolumn{5}{l}{\textit{Cartesian wrench}}\\
Avg RMSE (Ext-GT) & 4.1586 & 4.3910 & \textbf{3.8785} & 4.3054\\
Avg RMSE (Int-GT) & 4.6837 & 5.3108 & \textbf{4.4357} & 5.0918\\
\bottomrule
\end{tabular}
\end{table}
\section{Conclusion}\label{sec:conclusion}
We introduced \ac{kvark}, a sensorless force observer that couples a probabilistic residual-torque model with a variance-aware disturbance Kalman filter. By (i) learning residual means and input-dependent variance from excitation data and (ii) adapting process noise via \ac{vb}, the observer robustly down-weights uncertain virtual measurements and tracks time-varying disturbances. Across two experimental settings (free vs. loaded trajectories and contact with an F/T-instrumented tool), \ac{kvark} reduced joint-space and Cartesian wrench errors relative to \ac{gp}- and \ac{nn}-based baselines while preserving low latency.

{Limitations include evaluation on a single robot and limited analysis near zero velocities. In the surfacing task, Cartesian wrench estimation accuracy remains limited, particularly in tangential directions, due to contact-related effects discussed above. We also currently use \ac{kmp} predictive variance as a heteroscedastic confidence term without explicitly separating aleatoric and epistemic uncertainty. Future work will include multi-robot validation, temperature-dependent friction drift, tighter treatment of static friction/stick–slip, and an explicit uncertainty decomposition to support variance-aware safety thresholds in impedance control.}

The principled uncertainty quantification framework demonstrated in \ac{kvark} establishes a new paradigm for sensorless force estimation that advances both theoretical understanding and practical capabilities for next-generation collaborative robotics and human-robot interaction systems.

\bibliographystyle{IEEEtran}  
\bibliography{bib.bib}

\end{document}